%% file: main-manuscript.tex
\documentclass[manuscript,screen,nonacm]{acmart} 

\AtBeginDocument{%
  \providecommand\BibTeX{{%
    \normalfont B\kern-0.5em{\scshape i\kern-0.25em b}\kern-0.8em\TeX}}}

\usepackage{subcaption}

\usepackage{pdfpages}

\usepackage{ragged2e}
\usepackage{tabularx}
\usepackage{multirow}
\usepackage{float}

\def\corpus{MILPaC}
\def\IPdataset{MILPaC-IP}
\def\Actsdataset{MILPaC-Acts}
\def\CCIdataset{MILPaC-CCI-FAQ}

\setcopyright{none}
\renewcommand\footnotetextcopyrightpermission[1]{}

\begin{document}



\title{MILPaC: A Novel Benchmark for Evaluating Translation of Legal Text to Indian Languages}
\titlenote{This work is to be published in \textit{ACM Transactions on Asian and Low-Resource Language Information Processing (TALLIP)}}

\author{Sayan Mahapatra}
\authornote{The first two authors have contributed equally. Corresponding author: Debtanu Datta (\texttt{debtanudatta04@gmail.com})
}
\email{sayan.mahapatra.iitkgp@gmail.com}
\author{Debtanu Datta}
\authornotemark[2]
\email{debtanudatta04@gmail.com}
\author{Shubham Soni}
\email{shubhamsonikgp@gmail.com}
\author{Adrijit Goswami}
\email{goswami@maths.iitkgp.ac.in}
\author{Saptarshi Ghosh}
\email{saptarshi@cse.iitkgp.ac.in}
\affiliation{%
  \institution{Indian Institute of
Technology Kharagpur}
  \country{India}
}






\begin{abstract}
Most legal text in the Indian judiciary is written in complex English due to historical reasons. However, only a small fraction of the Indian population is comfortable in reading English. Hence legal text needs to be made available in various Indian languages, possibly by translating the available legal text from English. 
Though there has been a lot of research on translation to and between Indian languages, to our knowledge, there has not been much prior work on such translation in the legal domain. 
In this work, we construct the first high-quality legal parallel corpus containing aligned text units in English and nine Indian languages, that includes several low-resource languages. 
We also benchmark the performance of a wide variety of Machine Translation (MT) systems over this corpus, including commercial MT systems, open-source MT systems and Large Language Models. 
Through a comprehensive survey by Law practitioners, we check how satisfied they are with the translations by some of these MT systems, and how well automatic MT evaluation metrics agree with the opinions of Law practitioners.
\end{abstract}




\maketitle

\input{intro}

\input{related}

\input{datasets}

\input{benchmark}

\input{conclu}

\begin{acks}
We thank the Law students from the Rajiv Gandhi School of Intellectual Property Law, India -- Purushothaman R, Samina Khanum, Krishna Das, Md. Ajmal Ibrahimi, Shriyash Shingare -- who helped us in evaluating the translations and provided valuable suggestions and observations during the evaluation.
We thank Prof. Shouvik Guha and Prof. Anirban Mazumder of The West Bengal National University of Juridical Sciences, Kolkata for allowing us to use their ``My IP'' Primers.  
We are grateful to the Ministry of Law and Justice, Government of India, particularly Dr. Nirmala Krishnamoorthy, for allowing us to use the Central Acts in regional languages for our research.
We also thank the Competition Commission of India, particularly Mr. Pramod Kumar Pramod, for allowing us to use their FAQ booklets for developing our dataset.

Debtanu Datta is supported by the Prime Minister’s Research Fellowship (PMRF) from the Government of India.
The work is partially supported by the Technology Innovation Hub (TIH) on AI for Interdisciplinary Cyber-Physical Systems (AI4ICPS) set up by IIT Kharagpur under the aegis of DST, Government of India (GoI).
\end{acks}



\clearpage
\noindent {\bf \Large Appendix}

\appendix
\input{appendix}

\end{document}

%% file: intro.tex
\section{Introduction} \label{sec:intro}

Most legal text in the Indian judiciary is written in English due to historical reasons, and legal English is particularly complex.
However, only about 10\% of the general Indian population is comfortable in reading English.\footnote{\url{https://censusindia.gov.in/nada/index.php/catalog/42561}} 
Additionally, since different states in India have different local languages, a person residing in a non-native state may need to understand legal text written in a language with which the person is not familiar.
For these reasons, when the common Indian confronts a legal situation, this language barrier is a frequent problem. 
Thus, it is important that legal text be made available in various Indian languages, to improve access to justice for a very large fraction of the Indian population (especially the financially weaker sections). 
The Indian government is also promoting the need for legal education in various Indian languages.\footnote{\url{https://pib.gov.in/PressReleaseIframePage.aspx?PRID=1882225}} 

Since most legal text (e.g., laws of the land, manuals for legal education) is already available in English, the most practical way is to translate legal text from English to Indian languages. 
For this translation, one of the several existing Machine Translation (MT) systems can be used, including commercial systems such as the Google Translation system\footnote{\url{https://cloud.google.com/translate/docs/samples/translate-v3-translate-text}}, Microsoft Azure Translation API\footnote{\url{https://azure.microsoft.com/en-us/products/cognitive-services/translator}}, and so on.  
Also there has been a lot of academic research on developing MT systems to and between Indian languages~\citep{ramesh2021samanantar,siripragada-etal-2020-multilingual, kunchukuttan2020iwnparallel, https://doi.org/10.48550/arxiv.2001.09907}. 
However, to our knowledge, very few prior works have focused on Indian language translations \textit{in the legal domain} (see Section~\ref{sec:related} for a description of these works).
Hence, with respect to translation in the legal domain, some important questions that need to be answered are -- 
(i)~{\it how good are the existing Machine Translation (MT) systems for translating legal text to Indian languages?} and 
(ii)~{\it how well do standard MT performance metrics agree with the opinion of Law practitioners regarding the quality of translated legal text?}
Till date, there has not been any attempt to answer the questions stated above, and there does not exist any benchmark for evaluating the quality of the translation of legal text to Indian languages. 
\noindent In this work, we bridge this gap by making the following contributions:\\
$\bullet$ We develop \textbf{\corpus{}} (\textbf{M}ultilingual \textbf{I}ndian \textbf{L}egal \textbf{Pa}rallel \textbf{C}orpus), one of the first parallel corpus in Indian languages in the legal domain. 
The corpus consists of three high-quality datasets carefully compiled from reliable sources of legal information in India, containing parallel aligned text units in English and various Indian languages, including both Indo-Aryan languages (Hindi, Bengali, Marathi, Punjabi, Gujarati, \& Oriya) and Dravidian languages (Tamil, Telugu, \& Malayalam).\footnote{In this paper, we collectively refer to these languages as \textit{Indian languages}.} Note that many of these are low-resource languages. 
\corpus{} can be used for evaluating MT performance in translating legal text from English to various Indian languages, or from one Indian language to another. 
Different from existing parallel corpora in the Indian legal domain (see Section~\ref{sec:related}),  \corpus{}  is carefully curated from verified legal sources and ratified by Law experts, making it far more reliable for evaluating MT systems in the Indian legal domain.  Moreover, \corpus{}'s versatility allows its use in other tasks such as cross-lingual question answering, further distinguishing it from other datasets.\\
$\bullet$ We benchmark the performance of several MT systems over \corpus{}, including commercial systems, systems developed by the academic community, and the recently developed Large Language Models. 
The best performance is found for two commercial systems by Google and Microsoft Azure, and an academic translation system IndicTrans~\citep{ramesh2021samanantar}.\\
$\bullet$ A comprehensive survey is also carried out using Law students to understand how satisfied they are with the outputs of current MT systems and what errors are frequently committed by these systems.\\
$\bullet$ We also draw insights into the agreement of automatic translation evaluation metrics (such as BLEU, GLEU, and chrF++) with the opinion of Law students. We find mostly positive but low correlation of automatic metrics with the scores given by Law students for most Indian languages, although high correlation is observed for English-to-Hindi translations.

\begin{figure}[!htb]
\centering
\captionsetup{width=\textwidth}
\includegraphics[height=9cm, width=\textwidth]{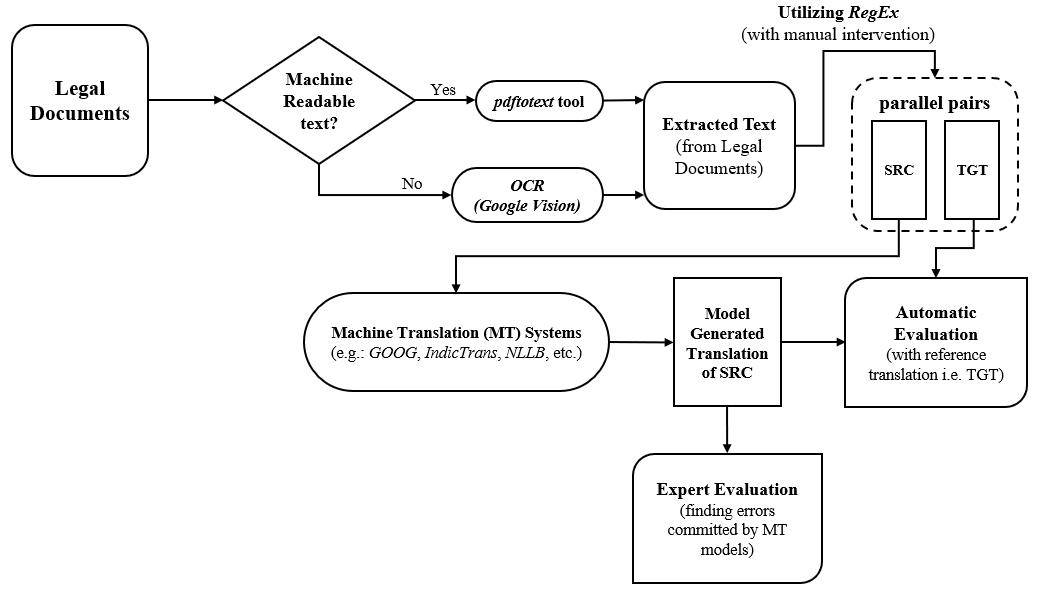}
\caption{A flow chart describing our methodology on benchmarking MT systems for legal text translation in Indian languages.}
\label{fig:flow_chart_process}
\end{figure}

A flowchart describing the methodology followed in this work is given in Figure~\ref{fig:flow_chart_process}.
The \corpus{} corpus is publicly available at \url{https://github.com/Law-AI/MILPaC} under a standard non-commercial license.
We believe that this corpus, as well as the insights derived in this paper, is an important step towards improving machine translation of legal text to Indian languages, a goal that is important for making Law accessible to a very large fraction of the Indian population.

%% file: related.tex
\section{Related Works} \label{sec:related}

\noindent \textbf{Parallel corpora for Indian and non-Indian languages (for non-legal domains):} 
Many parallel corpora exist for translations in the Indian languages. 
\citet{ramesh2021samanantar} released \textit{Samanantar}, the largest publicly available parallel corpora for 11 Indian languages. They also released IndicTrans, a transformer-based MT model trained on this dataset. 
Another parallel corpus for 10 Indian languages was released by~\citet{siripragada-etal-2020-multilingual}.
The \textit{FLORES-200} corpora for evaluation of MT in resource-poor languages was released by Facebook~\citep{NLLB}.
\textit{IndoWordNet corpus}~\citep{kunchukuttan2020iwnparallel} 
and \textit{PMIndia corpus}~\citep{https://doi.org/10.48550/arxiv.2001.09907} are other parallel corpora for Indian languages.
Also, there are several bilingual English-Indic corpora available, such as the \textit{IITB English-Hindi corpus}~\citep{kunchukuttan-etal-2018-iit}, the \textit{BUET English-Bangla corpus}~\citep{hasan-etal-2020-low}, the English-Tamil~\citep{biblio:RaBoMorphologicalProcessing2012} and English-Odia~\citep{parida2020odiencorp} corpus. 
Recently, \citet{Mizo_paper} created the first Mizo–English parallel corpus which covers the low-resource language Mizo.

There are several non-legal parallel corpora consisting of low-resource non-Indian languages as well, such as English-Kurmanji and English-Sorani \citep{Kurdish_Paper} bilingual corpora, \textit{KreolMorisienMT} \citep{KreolMorisienMT_Paper} which consists of two low-resource parallel corpora English-KreolMorisien and French-KreolMorisien, and \textit{ChrEn} \citep{ChrEn_Paper} which covers the endangered language Cherokee. Also, a new English-Arabic parallel corpus has been recently introduced by \citet{Arabic_Paper} to deal with phishing emails.
The above-mentioned works are outlined in Table~\ref{tab:non-legal-corpora}.
Note that \textit{none of the parallel corpora stated above are in the legal domain}.


\begin{table}[tb]
\small
\centering
\captionsetup{width=\textwidth}
\caption{Existing Parallel Corpora for Indian and Non-Indian Languages, in Non-Legal domains}
\begin{tabular}{|m{5.8cm}|m{2.7cm}|m{5.2cm}|}
\hline
\textbf{Corpus} & \textbf{Languages} & \textbf{Description} \\ \hline
\textit{Samanantar} \citep{ramesh2021samanantar} & 11 Indian languages & Largest publicly available parallel corpora for Indian languages \\ \hline
\textit{Parallel corpora} by~\citet{siripragada-etal-2020-multilingual} & 10 Indian languages & Sentence aligned parallel corpora \\ \hline
\textit{IndoWordNet corpus}~\citep{kunchukuttan2020iwnparallel} & 18 Indian languages & Contains about 6.3 million parallel segments \\ \hline
\textit{PMIndia corpus}~\citep{https://doi.org/10.48550/arxiv.2001.09907} & 13 Indian languages & Corpus includes up to 56K sentences for each language pair \\ \hline
\textit{IITB English-Hindi corpus}~\citep{kunchukuttan-etal-2018-iit} & English-Hindi & Bilingual corpus for MT task \\ \hline
\textit{BUET English-Bangla corpus}~\citep{hasan-etal-2020-low} & English-Bangla & Bilingual corpus for MT task \\ \hline
\textit{English-Tamil corpus}~\citep{biblio:RaBoMorphologicalProcessing2012} & English-Tamil & Bilingual corpus for MT task \\ \hline
\textit{English-Odia corpus}~\citep{parida2020odiencorp} & English-Odia & Bilingual corpus for MT task \\ \hline
\textit{English-Mizo corpus}~\citep{Mizo_paper} & English-Mizo & First parallel corpus for the low-resource language Mizo \\ \hline
\textit{FLORES-200}~\citep{NLLB} & 200 languages & Corpora that includes many resource-poor languages \\ \hline
\textit{Kurdish parallel corpus}~\citep{Kurdish_Paper} & Sorani-Kurmanji, English-Kurmanji, English-Sorani & Parallel corpora for the two major dialects of Kurdish -- Sorani and Kurmanji \\ \hline
\textit{KreolMorisienMT}~\citep{KreolMorisienMT_Paper} & English-KreolMorisien, French-KreolMorisien & Parallel corpora for low-resource KreolMorisien language \\ \hline
\textit{ChrEn corpus}~\citep{ChrEn_Paper} & English-Cherokee & Covers the endangered language Cherokee \\ \hline
\textit{English-Arabic corpus}~\citep{Arabic_Paper} & English-Arabic & Parallel corpus for phishing emails detection \\ \hline
\end{tabular}
\label{tab:non-legal-corpora}
\end{table}

\noindent \textbf{Parallel corpora in the legal domain for non-Indian languages:}
Organizations such as the United Nations (UN) and the European Union (EU) have developed several parallel corpora in the legal domain for MT.
The \textit{UN Parallel Corpus v1.0} contains parliamentary documents and their translations over 25 years, 
in six official UN-languages~\citep{ziemski-etal-2016-united}.
Again, based on the European parliament proceedings, \citet{koehn-2005-europarl} released \textit{Europarl}, a parallel corpus in 11 official languages of the EU. 
\citet{hofler2014constructing} released a multilingual corpus - \textit{Bilingwis Swiss Law Text collection} based on the classified compilation of Swiss federal legislations.
The EU also released the \textit{EUR-Lex parallel corpora} including multiple parallel corpora across 24 European languages~\citep{baisa2016european}. 
These works are summarized in Table~\ref{tab:legal-non-indian-corpora}.
In comparison, there has only been little effort in the Indian legal domain towards the development of parallel corpora for translation.


\begin{table}[tb]
\small
\centering
\captionsetup{width=\textwidth}
\caption{Existing Parallel Corpora in the Legal Domain for Non-Indian Languages}
\begin{tabular}{|m{5.8cm}|m{2.7cm}|m{5.2cm}|}
\hline
\textbf{Corpus} & \textbf{Languages} & \textbf{Description} \\ \hline
\textit{UN Parallel Corpus v1.0}~\citep{ziemski-etal-2016-united} & 6 official UN languages & Parliamentary documents and their translations over 25 years \\ \hline
\textit{Europarl}~\citep{koehn-2005-europarl} & 11 official EU languages & Based on the European parliament proceedings \\ \hline
\textit{Bilingwis Swiss Law corpus}~\citep{hofler2014constructing} & 4 official Swiss languages & Based on the classified compilation of Swiss federal legislations \\ \hline
\textit{EUR-Lex parallel corpus}~\citep{baisa2016european} & 24 European languages & Largest parallel corpus built from European Legal documents \\ \hline
\end{tabular}
\label{tab:legal-non-indian-corpora}
\end{table}

\begin{table}[!htb]
\small
\centering
\captionsetup{width=\textwidth}
\caption{Existing Parallel Corpora in the Legal Domain for Indian Languages}
\begin{tabular}{|m{5.8cm}|m{2.7cm}|m{5.2cm}|}
\hline
\textbf{Corpus} & \textbf{Languages} & \textbf{Description} \\ \hline
\textit{LTRC Hindi-Telegu parallel corpus}~\citep{mujadia-sharma-2022-ltrc} & Hindi-Telugu & Small parallel corpus for legal text, only in the 2 Indian languages Hindi and Telegu \\ \hline
\textit{Anuvaad parallel corpus} & 9 Indian languages & Synthetic parallel corpus, suitable for training but not for evaluating since their mappings can be inaccurate \\ \hline
\corpus{} (\textit{developed in this work}) & 9 Indian languages & High-quality corpus curated from verified legal sources; suitable for evaluating and benchmarking MT systems; can also be utilized for cross-lingual question answering
\\ \hline
\end{tabular}
\label{tab:legal-indian-corpora}
\end{table}

\noindent \textbf{Parallel corpora in the legal domain for Indian languages:}
To our knowledge, there are only two parallel corpora for the Indian legal domain (summarized in Table~\ref{tab:legal-indian-corpora}).
First, the \textit{LTRC Hindi-Telegu parallel corpus}~\citep{mujadia-sharma-2022-ltrc} contains a small parallel corpus from the legal domain, but only in the 2 Indian languages Hindi and Telegu (whereas the \corpus{} corpus developed in this work covers as many as 9 Indian languages). 
Second, the \textit{Anuvaad}\footnote{\url{https://github.com/project-anuvaad/anuvaad-parallel-corpus}} parallel corpus for legal text in 9 Indian languages.
But the corpus seems synthetically created, i.e., the texts of different languages are mapped automatically. Hence, this corpus can be used for training MT models but is not reliable for evaluating MT models due to potential inaccuracies in mappings.
Moreover, the legal data sources are not well documented, and there is no detail regarding the quality / expert evaluation of the corpus. Also, \textit{Anuvaad} does not have the Oriya language. In contrast, the \corpus{} corpus is carefully curated from verified legal sources and ratified by Law experts, making it far more reliable and suitable for evaluating MT systems in the Indian legal domain. Also, \corpus{} covers Oriya. Moreover, \corpus{}'s versatility extends to tasks such as cross-lingual question answering, further distinguishing it from other datasets.



%% file: datasets.tex
\section{The \corpus{} Datasets} \label{sec:datasets}

This section describes 
the \textbf{Multilingual Indian Legal Parallel Corpus ({\corpus})} which comprises of 3 datasets, comprising a total of 17,853 parallel text pairs across English and 9 Indian languages.

\subsection{Dataset 1: {\IPdataset}}
We develop this dataset from a set of primers released by a society of Law practitioners -- the WBNUJS Intellectual Property and Technology Laws Society (IPTLS) -- from a reputed Law school in India. 
The primers are released in English~(EN) and 9 Indian languages, \textit{viz.}, Bengali~(BN), Hindi~(HI), Marathi~(MR), Tamil~(TA), Gujarati~(GU), Telugu~(TE), Malayalam~(ML), Punjabi~(PA), Oriya~(OR). 
Each primer contains approximately 57 question-answer (QA) pairs related to Indian Intellectual Property laws, each QA pair given in all the Indian languages stated above.

The 10 PDF documents (one corresponding to the primer in each language) were downloaded from the website of the IPTLS society stated above.\footnote{\url{https://nujsiplaw.wordpress.com/my-ip-a-series-of-ip-awareness-primers/}} 
The PDF documents had embedded text which was extracted using the \textit{pdftotext}\footnote{\url{https://www.xpdfreader.com/pdftotext-man.html}}
utility. 
Regular expressions were used to identify the start of each question and answer. 
Each QA pair in these primers was numbered consistently across all languages, and we used this numbering to identify parallel textual units.

\begin{table}[tb]
\centering
\captionsetup{width=\textwidth}
\caption{Number of parallel text units per language pair in (1)~{\IPdataset} - black entries in upper triangular part, and (2)~{\CCIdataset} - blue italicized entries in lower triangular part. For both datasets, text units are QA-pairs, hence not tokenized into sentences (details in text). 
The language codes (used throughout  the paper) are: English (EN), Bengali (BN), Hindi (HI), Marathi (MR), Tamil (TA), Telugu (TE), Malayalam (ML), Punjabi (PA), Oriya (OR), Gujarati (GU).}
\begin{tabular}{|c|cccccccccc|} 
 \hline
  & \textbf{EN} & \textbf{BN} & \textbf{HI} & \textbf{MR} & \textbf{TA} & \textbf{TE} & \textbf{ML} & \textbf{PA} & \textbf{OR} & \textbf{GU} \\ 
 \hline
 \textbf{EN} & $\times$ & 110 & 114 & 114 & 114 & 112 & 114 & 114 & 114 & 114 \\ 
 \textbf{BN} & \textcolor{blue}{\it 365} & $\times$ & 110 & 110 & 110 & 108 & 110 & 110 & 110 & 110 \\
 \textbf{HI} & \textcolor{blue}{\it 365}  & \textcolor{blue}{\it 365} & $\times$ & 114 & 114 & 112 & 114 & 114 & 114 & 114 \\ 
 \textbf{MR} & \textcolor{blue}{\it 365} & \textcolor{blue}{\it 365}  &  \textcolor{blue}{\it 365}& $\times$ & 114 & 112 & 114 & 114 & 114 & 114 \\
 \textbf{TA} & \textcolor{blue}{\it 365} & \textcolor{blue}{\it 365} & \textcolor{blue}{\it 365} & \textcolor{blue}{\it 365} & $\times$ & 112 & 114 & 114 & 114 & 114 \\
 \textbf{TE} &  &  &  &  &  & $\times$ & 112 & 112 & 112 & 112 \\
 \textbf{ML} &  &  &  &  &  &  & $\times$ & 114 & 114 & 114 \\
 \textbf{PA} &  &  &  &  &  &  &  & $\times$ & 114 & 114 \\
 \textbf{OR} &  &  &  &  &  &  &  &  & $\times$ & 114 \\
 \textbf{GU} &  &  &  &  &  &  &  &  &  & $\times$ \\
 \hline
\end{tabular}
\label{tab:MILPaC-IP-CCI-Counts}
\end{table}

Each question and each answer is considered as a `textual unit' in this dataset.
Note that, we decided to {\it not} break the questions/answers into sentences, due to the following reasons: 
First, we have alignments in terms of questions and answers, and not their individual sentences. If we split each question and answer into sentences, aligning the sentences would require synthetic methods (like nearest neighbour search) which might have introduced some inaccuracies in the dataset. 
Since the primers are a high-quality parallel corpus developed manually by Law practitioners, we decided not to use such synthetic methods to maintain a high standard of the parallel corpus.
Second, since this dataset contains QA pairs, it can be used for other NLP tasks such as Cross-Lingual Question-Answering. Tokenizing the text units into sentences would remove this potential utility of the dataset. 

The number of parallel pairs in the \IPdataset{} dataset for various language-pairs is given in Table~\ref{tab:MILPaC-IP-CCI-Counts} (black entries in the upper triangular part).
The slight variations in the entries for different language-pairs in Table~\ref{tab:MILPaC-IP-CCI-Counts} are explained in Appendix~\ref{sub:appendix-ipdataset-variation}.
Some examples of text pairs of the \IPdataset{} are given in Figure~\ref{fig:Legal-Indic-IP-Sample}.

\subsection{Dataset 2: {\CCIdataset}}


The second parallel corpus is developed from a set of FAQ booklets released by the Competition Commission of India (CCI). 
These FAQ booklets are based on {\bf The Competition Act, 2002} statute that provides the legal framework to deal with competition issues in India.
The FAQs are written in English and 4 Indian languages -- Bengali, Hindi, Marathi, and Tamil. 
Each FAQ booklet contains 184 QA pairs.
Each FAQ booklet has been manually translated by CCI officials and can be treated as expert translation, which makes this a reliable source for authentic multilingual parallel data.

The FAQ booklets (PDF documents) corresponding to each language were downloaded from the CCI website\footnote{\url{https://www.cci.gov.in/advocacy/publications/advocacy-booklets}}. 
These documents are not machine readable; hence, the text was extracted using OCR (as done by~\citet{ramesh2021samanantar}).
We tried out two OCR systems - Tesseract\footnote{\url{https://github.com/tesseract-ocr/tesseract}} and Google Vision API\footnote{\url{https://cloud.google.com/vision/docs/samples/vision-document-text-tutorial}}. 
Both OCR systems support English and all 9 Indian languages considered in this work.
An internal evaluation (details in Appendix, Section~\ref{OCR Details}) showed that the Google Vision had better text detection accuracy than Tesseract.
Hence, we decided to use Google Vision. 

After applying OCR to the PDF documents, a single text file is obtained, which is annotated to separate and align the text units.
For all QA pairs, the translations had the same question number in all the documents, and this information was used to align the QA pairs.
Note that, as in {\IPdataset}, we do \textit{not} split the questions and answers into sentences for the same reasons as discussed above. 

\begin{figure}[htb]
    \centering
    \begin{subfigure}[b]{0.8\textwidth}
        \centering
        \includegraphics[width=\textwidth, height=5cm]{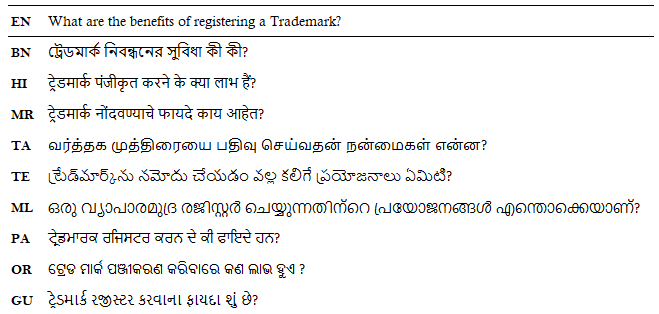}
        \caption{Examples of parallel pairs in {\IPdataset}.}
        \label{fig:Legal-Indic-IP-Sample}
    \end{subfigure}

    \begin{subfigure}[b]{0.8\textwidth}
        \centering
        \includegraphics[width=\textwidth, height=3.5cm]{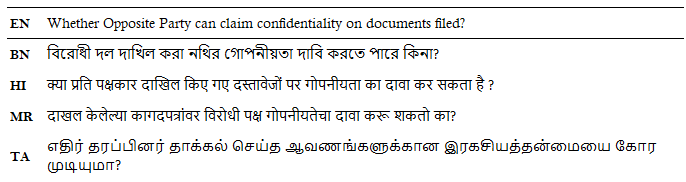}
        \caption{Examples of parallel pairs in {\CCIdataset}.}
        \label{fig:MILPaC-CCI-FAQ-Samples}
    \end{subfigure}

    \begin{subfigure}[b]{0.8\textwidth}
        \centering
        \includegraphics[width=\textwidth, height=5.5cm]{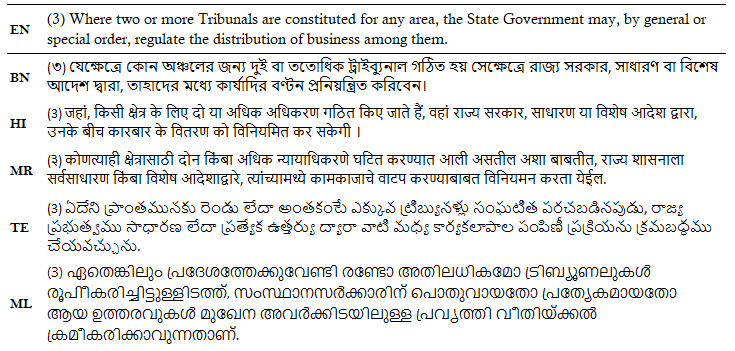}
        \caption{Examples of parallel pairs in {\Actsdataset}.}
        \label{fig:MILPaC-ACTS-Samples}
    \end{subfigure}

    \caption{Examples of parallel text pairs from the 3 datasets in {\corpus}. From each dataset, we show one textual unit in all available languages in that dataset.}
    \label{fig:MILPaC-samples}
\end{figure}

The number of parallel text pairs for various language-pairs in the \CCIdataset{} dataset are given in the Table~\ref{tab:MILPaC-IP-CCI-Counts} (blue italicized entries in lower triangular part).
As stated earlier, each FAQ booklet contains 184 QA pairs. The answers for 3 QA pairs are given in tables; for these, we did not include the answers and considered only the questions, because of difficulties in accurate OCR of tables. Hence, for each language-pair, we have $184 \times 2 - 3 = 365$ text pairs, as stated in Table~\ref{tab:MILPaC-IP-CCI-Counts}.
Some example text pairs are given in Figure~\ref{fig:MILPaC-CCI-FAQ-Samples}.


 



\subsection{Dataset 3: {\Actsdataset}}

This dataset is made up from PDF scans (non-machine readable) of Indian Acts (statutes or written laws) documents published by the Indian Legislature.
Unlike the previous two datasets (which are already compiled in multiple languages), this parallel corpus is {\it not} available readily, and is very challenging and time-consuming to develop. 

The first challenge is that, while a lot of Indian Acts are available in English and Hindi, very few are available in other Indian languages.
Even among the few Acts that are available in other Indian languages, only a small fraction are suitable for OCR (most other PDFs are of poor texture and resolution). 
The challenges in OCR of Indian Legal Acts are described in the Appendix (section~\ref{sec:OCR-Challenges}).

The English and translated versions of the Acts were obtained from the website of the Indian legislature\footnote{\url{https://legislative.gov.in/central-acts-in-regional-language/}}.
Candidate Acts were selected based on two criteria -- (i)~\textit{availability of translations in multiple languages} (so as to get translation-pairs across Indian-Indian languages as well), and \textit{availability of good quality scans}.
10 Acts were selected as per these criteria;
the selected Acts are listed in Appendix, Section~\ref{sec:appendix-acts}.

\begin{table}[tb]
\centering
\captionsetup{width=\textwidth}
\caption{Number of Parallel Text units per language pair in {\Actsdataset}. Text units are tokenized into sentences for this dataset.}
\begin{tabular}{|c|cccccccccc|} 
 \hline
   & \textbf{EN} & \textbf{BN} & \textbf{HI} & \textbf{MR} & \textbf{TA} & \textbf{TE} & \textbf{ML} & \textbf{PA} & \textbf{OR} & \textbf{GU}\\
 \hline
 \textbf{EN} & $\times$ & 739  & 706 & 578 & 418 & 319 & 443 & 261 & 256 & 316\\ 
 \textbf{BN} &  & $\times$  & 439 & 439 & $\times$ & 319 & 438 & $\times$ & $\times$ & $\times$\\
 \textbf{HI} &  &  & $\times$ & 578 & $\times$ & 319 & 443 & 262 & 256 & $\times$\\ 
 \textbf{MR} &  &  &  & $\times$ & $\times$ & 319 & 443 & 133 & 128 & $\times$\\
 \textbf{TA} &  &  &  &  & $\times$ & $\times$ & $\times$ & $\times$ & $\times$ & $\times$\\
 \textbf{TE} &  &  &  &  &  & $\times$ & 319 & $\times$  & $\times$ & $\times$\\
 \textbf{ML} &  &  &  &  &  &  &  $\times$ & $\times$  & $\times$ & $\times$\\
 \textbf{PA} &  &  &  &  &  &  &   & $\times$ & 256 & $\times$\\
 \textbf{OR} &  &  &  &  &  &  &  &  & $\times$ & $\times$\\
 \textbf{GU} &  &  &  &  &  &  &  &  &  & $\times$\\
 \hline
\end{tabular}
\label{tab:MILPaC-Acts-Counts}
\end{table}

As discussed previously for the \CCIdataset{} dataset, Google Vision was chosen for OCR.
A similar process was followed to get one text file corresponding to each Act, and the extracted text was then annotated using the same semi-automatic approach as was described for {\CCIdataset}.
We identified parallel text units (paragraphs, bulleted list elements) by looking at the versions in the two languages and using structural information such as paragraphs, and section/clause numbering.
Table~\ref{tab:MILPaC-Acts-Counts} gives the number of parallel text pairs for different language-pairs in \Actsdataset{}.
Some example text pairs of \Actsdataset{} are given in Figure~\ref{fig:MILPaC-ACTS-Samples}.



Note that, compared to {\IPdataset} and {\CCIdataset} datasets, aligning the Acts documents is much more challenging and requires frequent manual intervention.
The documents are long, and often have multi-column text layout, which confuses OCR systems leading to text segmentation issues (see Appendix, Section~\ref{sec:OCR-Challenges} for the main challenges in OCR). 
Such factors frequently lead to OCR errors that need to be manually corrected by rearranging the extracted text by referring back to the source document. 
Hence developing this dataset is very time-consuming. This is why, till date, we have been able to process only 10 Acts. 
Even after the manual corrections, a fraction of the data had to be discarded since they could not be aligned across languages. 
This non-alignment was mainly due to the factors that are explained in Appendix Section~\ref{sec:OCR-Challenges}.
According to our estimate, in the \Actsdataset{}, approximately 10\% of the textual units had to be discarded on average across all the Acts we considered. The percentage of text units discarded for individual Acts varies between 8\% and 14\%. 
All these steps have been taken to ensure the high quality of the dataset, making it very time-consuming to develop.

\subsection{Comparing the three datasets in \corpus{}}
\noindent \textbf{Length of textual units:} 
The distribution of the number of tokens in the textual units of the three datasets is shown in Table~\ref{tab:Length-Table}.
The Moses tokenizer\footnote{\url{https://github.com/alvations/sacremoses}} has been used for tokenizing English sentences, and the IndicNLP library\footnote{\url{https://github.com/anoopkunchukuttan/indic_nlp_library}} has been used for tokenizing sentences in Indian languages.
Across all three datasets, most of the textual units are shorter than 20 tokens. 
However, the CCI-FAQ dataset has 22\% textual units having more than 100 tokens.

\begin{table}[ht]
\centering
\captionsetup{width=\textwidth}
\caption{Length distribution of text units in {\corpus}}
\begin{tabular}{|cccc|} 
\hline
\multirow{2}{*}{\# tokens}  & \multicolumn{3}{c|}{\% of text units} \\
& {\IPdataset} & {\CCIdataset} & {\Actsdataset} \\
\hline
1-10   & 34.5\% & 17.6\% & 18.4\% \\
11-20  & 16.0\% & 22.0\% & 23.7\%\\
21-30  & 4.7\%  &  9.4\% & 19.6\% \\
31-40  & 6.2\%  &  5.8\% & 13.2\%\\
41-50  & 5.9\%  &  4.9\% & 8.8\%\\
51-60  & 6.3\%  &  4.9\% & 5.7\% \\
61-70  & 5.6\%  &  4.7\% & 3.5\% \\
71-80  & 4.5\%  &  3.0\% & 2.3\% \\
81-90  & 4.0\%  &  3.7\% & 1.4\%\\
91-100 & 3.2\%  &  1.8\% & 1.0\%\\
$>$ 100  & 9.3\%  & 22.2\% & 2.2\%\\
\hline
\end{tabular}
\label{tab:Length-Table}
\end{table}

\noindent \textbf{Text complexity of the datasets:}
To compare the three datasets in terms of the complexity of the language contained in them, we have computed two well-known text readability metrics -- \textit{The Flesch Kincaid Grade Level} (FKGL) \& \textit{Gunning Fog Index} (GFI). Higher values of these metrics indicate that the text is more complex, and requires higher levels of expertise/proficiency to comprehend. 
The results are shown in Table~\ref{tab:readability-table}.
Both metrics imply that the \Actsdataset{} dataset is the most complex (suitable only for highly educated readers, more complex than a standard research paper), the \CCIdataset{} dataset has intermediate complexity (suitable for college students), and the \IPdataset{} dataset is least complex (suitable for high school students). 
These differences are expected since the Acts (out of which \Actsdataset{} is constructed) are the most formal versions of the laws of the land which need to be precise and are hence stated using very formal language. 
Whereas the other two collections, particularly \IPdataset{}, are meant to be more simple and easily comprehensible for various types of stakeholders, including the general public. Hence, CCI-FAQ and IP datasets are in the form of simplified questions and answers. 

\begin{table}[tb]
\centering
\captionsetup{width=\textwidth}
\caption{Comparing the language complexity / readability of the three datasets in \corpus{} using two popular readability metrics.}
\begin{tabular}{|l|c|c|}
     \hline
     Datasets & FKGL & GFI \\
     \hline
     \IPdataset{} & 8.3 & 8.9 \\
     \CCIdataset{} & 15.1 & 12.2 \\
     \Actsdataset{} & 20.4 & 18.7 \\
     \hline
\end{tabular}
\label{tab:readability-table}
\end{table}

\noindent \textbf{Original and Translation Languages:} For the \IPdataset{} dataset, it is specified that the English version is the original version, and the other versions are translated from English. 
For \CCIdataset{} and \Actsdataset{}, the original language is not clearly stated anywhere. But both these datasets are based on Indian legal statutes / Acts, which have been officially published in English and Hindi since Indian independence. 
Hence, we believe that the English and Hindi versions are the original ones, and the versions in the other languages are translated from one of these two.

\subsection{Quality of the datasets}
We have attempted to make {\corpus} a high-quality parallel corpus for evaluating MT systems, through the following steps.

\noindent \textbf{Reliability of the sources:} 
For {\Actsdataset} and {\CCIdataset}, the data was extracted from the official Government of India websites.
The data for {\IPdataset} is published by one of India's most reputed Law Schools.
Thus, the translations that we used for constructing the dataset have been done by Government officials/Law practitioners and hence are likely to be of high quality.

\noindent \textbf{Manual corrections:} We manually verified a large majority of the data extracted from the PDFs. Wherever the OCR accuracy was compromised due to the poor resolution/quality of the PDFs (notably in \Actsdataset), we manually corrected the alignments, thus ensuring the quality of the dataset.

\noindent \textbf{Evaluation of the corpus quality by Law students:} 
Finally, for assessing the quality of our datasets, we recruited 5 senior LLB and LLM students who are native speakers of  Hindi, Bengali, Tamil, \& Marathi and are also fluent in English, from a reputed Law school in India (details given later in Section~\ref{sub:expert-evaluation}). Hence this evaluation was carried out over these four languages.
We asked the Law students to read a large number of English text units and their translations (from the \corpus{} corpus) in their native Indian language (across all 3 datasets), and rate the quality of each text-pair on a scale of 1-5, with 1 being the lowest and 5 the highest. 
The purpose of the survey was explained to the Law students, and they were paid a mutually agreed honorarium for their annotations.
On average, 94\% of the parallel text-pairs achieved ratings of 4 or 5 from the Law students, which further demonstrates the high quality of the \corpus{} corpus. 


\subsection{Availability of the corpus}

The \corpus{} corpus is publicly available at \url{https://github.com/Law-AI/MILPaC} under a standard non-commercial license.
All the legal documents and translations used to develop the \corpus{} datasets are publicly available data on the Web. 
We have obtained formal permissions for using and publicly sharing the datasets -- 
from the Competition Commission of India (CCI) for the \CCIdataset{} dataset,
from the WBNUJS Intellectual Property and Technology Laws Society (IPTLS) for the \IPdataset{} dataset, and
from the Ministry of Law and Justice, Government of India for using and publicly sharing the \Actsdataset{} dataset.

%% file: benchmark.tex
\section{Benchmarking Translation Systems} \label{sec:benchmark}


We now use the \corpus{} corpus to benchmark the performance of established state-of-the-art Machine Translation (MT) systems that have performed well on other MT benchmark datasets. Though \corpus{} can be used to evaluate Indian-to-Indian translations as well, we primarily consider English-to-Indian translation since that is the most common use-case in the Indian legal domain.

\subsection{Translation systems}

We consider the following MT systems, some of which are commercial, while others are publicly available MT systems. These systems were selected because of their strong performances on well-known MT benchmarks such as FLORES-101~\citep{flores}, WAT2021\footnote{\url{https://lotus.kuee.kyoto-u.ac.jp/WAT/WAT2021/index.html}}, and WAT2020~\citep{WAT2020bib}.

\vspace{2mm}
\noindent $\bullet$ \textbf{Commercial MT Systems:}

\noindent
(1)~\underline{\textbf{GOOG}}: the Google Cloud Translation - Advanced Edition (v3) system\footnote{\url{https://cloud.google.com/translate/docs/samples/translate-v3-translate-text}}, which is one of the most popular and widely used commercial MT systems globally. We use the Translation API provided by the Google Cloud Platform (GCP) to access this system (which we abbreviate as GOOG).

\noindent
(2)~\underline{\textbf{MSFT}}: This is another well-known commercial MT system developed by Microsoft. Similar to GOOG, we use the Translation API offered by Microsoft Azure Cognitive Services (v3)\footnote{\url{https://azure.microsoft.com/en-us/products/cognitive-services/translator}} to access the system (which we abbreviate as MSFT). 

\noindent
(3)~\underline{\textbf{Large Language Models}}: We have also benchmarked the translation performances of two popular commercial Large Language Models (LLMs) in the GPT-3.5 family\footnote{\url{https://platform.openai.com/playground/p/default-translation}} developed by OpenAI -- \underline{\textbf{Davinci-003}} (formally called \textit{`text-davinci-003'}) and \underline{\textbf{GPT-3.5T-Inst}} (formally called \textit{`gpt-3.5-turbo-instruct'}). These are some of the most powerful models of OpenAI, with GPT-3.5T-Inst standing as the latest GPT-3.5-turbo model within the InstructGPT class. Similar to other commercial systems, we use the API provided by OpenAI to utilize these LLMs. 
Note that both these LLMs have a token limit of 4096 on the length of the input prompt plus the length of the generated output, where a token generally corresponds to $\sim4$ characters of common English text. 

\noindent
\textbf{Prompt for translation:} 
We conduct the translation task using these LLMs in an \textit{one-shot} configuration where one example translation is provided within the instruction-based prompt, before asking the LLM to translate an unseen text unit.\footnote{We tried both zero-shot and one-shot translation, and observed better results for one-shot. Hence we report the results of only the one-shot configuration.} 
Suppose we want to translate the textual unit $x_i$ from English to the target language <Target\_lang>. 
For each <Target\_lang> (Indian language), we identify another textual unit $e$ in English and its translation $t$ in <Target\_lang> from the gold standard translations in \corpus{}, to be given as an example within the prompt.
This example $e$ is chosen in consultation with the Law students (who participated in our surveys, details given later in Section~\ref{sub:expert-evaluation}), such that they are fully satisfied with its gold standard translations in all Indian languages which the Law practitioners know (Hindi, Bengali, Tamil, Marathi).
Once $e$ (in English) and $t$ (in <Target\_lang>) are selected, then we gave the prompt to the LLM as
``\textit{Translate from English to <Target\_lang>: $e => t~[\backslash n]~x_i =>~$}'' so that the model can learn from the given example translation ($e \rightarrow t$) and can then translate $x_i$ into the target language. Also, the text unit given as an example for one-shot translation, was not included in its evaluation.

\vspace{2mm}
\noindent
$\bullet$ \textbf{Open-source MT Systems:}

\noindent
(5)~\underline{\textbf{mBART-50}}: It is a transformer-based seq-to-seq model with multilingual fine-tuning (using the parallel data \textit{ML50}) from pre-trained mBART model~\citep{liu-etal-2020-multilingual-denoising}, that has 12 encoder layers and 12 decoder layers with model dimension of 1024 on 16 attention heads ($\sim$680M parameters). We use the fine-tuned checkpoint of the multilingual-BART-large-50 model\footnote{\url{https://huggingface.co/facebook/mbart-large-50-one-to-many-mmt}}~\citep{tang2020multilingual} for this translation task. This model can translate English to 7 of the Indian languages in our datasets. Oriya and Punjabi are not supported by mBART-50; hence mBART-50 is {\it not} evaluated for Oriya and Punjabi.

\noindent
(6)~\underline{\textbf{OPUS}}: It is another one-to-many transformer-based Neural Machine Translation (NMT) system\footnote{\url{https://huggingface.co/Helsinki-NLP/opus-mt-en-mul}}~\citep{Tiedemann2020OPUSMTB} that can translate from English to all 9 Indian languages in our \corpus{} corpus. It has a standard transformer architecture with 6 self-attentive layers in both the encoder and the decoder network with 8 attention heads in each layer and hidden state dimension of 512 ($\sim$74M parameters). It is trained over the large bitext repository \textit{OPUS}. 

\noindent
(7)~\underline{\textbf{NLLB}}: A seq-to-seq multilingual MT model~\citep{NLLB} which is also based on the transformer encoder-decoder architecture. Two variants of the NLLB are evaluated -- the \underline{\textbf{NLLB-1.3B}}\footnote{\url{https://huggingface.co/facebook/nllb-200-1.3B}} and \underline{\textbf{NLLB-3.3B}}\footnote{\url{https://huggingface.co/facebook/nllb-200-3.3B}} parameters models. 
NLLB-1.3B is a dense encoder-decoder model with $\sim$1.3B parameters with model dimension of 1024, 16 attention heads, 24 encoder layers and 24 decoder layers. NLLB-3.3B is another dense model with a larger model dimension of 2048, 16 attention heads, 24 encoder layers, and 24 decoder layers ($\sim$3.3B parameters). These variants are trained over \textit{NLLB-SEED} and \textit{PublicBitext}. 
NLLB is primarily intended for translation to low-resource languages. It has been reported to outperform several state-of-the-art MT models over the FLORES-101 benchmark, for many low-resource languages. It allows for translation among 200 languages, which include all the 9 Indian languages present in our corpus.

\noindent
(8)~\underline{\textbf{IndicTrans}}: A transformer-4x based multilingual NMT model\footnote{\url{https://github.com/AI4Bharat/indicTrans}} trained over the \textit{Samanantar} dataset for translation among Indian languages~\citep{ramesh2021samanantar}. IndicTrans has 6 encoder and 6 decoder layers, input embeddings of size 1536 with 16 attention heads ($\sim$434M parameters). It is a single-script model. All languages have been converted into the Devanagari script. As per the authors, this allows better lexical sharing between languages for transfer learning, while preventing fragmentation of the sub-word vocabulary between Indian languages and allows using a smaller subword vocabulary. 
Moreover, IndicTrans is specifically trained for translation between English and Indian languages, and for English-to-Indian and Indian-to-English translations. 
IndicTrans has been reported to outperform many other MT models over the established MT benchmarks like FLORES-101, WAT2021, and WAT2020.

There are 2 variants available for IndicTrans -- Indic-to-English and English-to-Indic which support 11 Indian languages, including all 9 of the Indian languages present in our corpus. The English-to-Indic model is used for our experiments. We refer the readers to the original paper~\citep{ramesh2021samanantar} for more details.

\vspace{2mm}
\noindent We use all the above-mentioned systems in their default settings, and without any fine-tuning, since we want to check their performance off-the-shelf.

\subsection{Metrics for automatic evaluation}

We use the following standard metrics to evaluate the performance of MT systems:

\noindent
$\bullet$ \textbf{BLEU}: BLEU (Bi-Lingual Evaluation Understudy) is an automatic MT evaluation metric~\citep{papineni-etal-2002-bleu}. It measures the overlap between the translation generated by an MT system ($t_h$) and the reference translation ($t_r$), by considering n-grams based precision ($n = 1, 2, 3, 4$). It calculates the highest occurrence of n-gram matches, and to avoid counting the same n-gram multiple times, it limits the count of matching n-gram to the highest frequency found in any single reference translation. 
This \textit{Clipped Count} is used for computing \textit{Clipped Precision} ($p_n$). Next, a term called `Brevity Penalty' (BP) is used to prevent the overfitting of sentence length and to penalize the generated translation that is too short. The formulas for BP and BLEU are as follows:
\begin{equation}\label{eq1}
    BP = \begin{cases}
    1, & \text{if } |t_h| > |t_r|\\
    \exp\left(1 - \frac{|t_r|}{|t_h|}\right), & \text{if } |t_h| \leq |t_r|
    \end{cases} 
\end{equation}
\begin{equation}\label{eq2}
    \begin{split}
        BLEU & = BP \cdot Geometric~Average~Precision \\
        & = BP \cdot \exp \left({\sum_{n=1}^{N}} w_n \log{p_n}\right)
    \end{split}
\end{equation}
where $|t_h|$ denotes the number of tokens in the generated translation, $|t_r|$ is the number of tokens in the reference translation, and $w_n$ represents the weight of the $n^{th}$ gram. In general, $N$ is 4.

\noindent
For calculating BLEU scores, we follow the approach of~\cite{ramesh2021samanantar} -- we first use the IndicNLP tokenizer to tokenize text in Indian languages. The tokenized text is then fed into the \textit{SacreBLEU} package~\citep{post-2018-call}. We state the \textit{SacreBLEU} signature\footnote{\textit{SacreBLEU} signature for BLEU: $BLEU+nrefs:1+case:mixed+eff:no+tok:none+smooth:exp+version:2.2.0$} to ensure uniformity and reproducibility across the models.

\noindent
$\bullet$ \textbf{GLEU}: 
GLEU (Google\_BLEU)~\citep{wu2016googles} is a modified metric inspired from BLEU and is known to be a more reliable metric for evaluation of translation of individual text units, since BLEU was designed to be a corpus-wide measure. 
In the GLEU measurement process, all subsequences comprising 1, 2, 3, or 4 tokens from both the generated and reference translations (referred to as n-grams) are recorded. Then, this n-grams based precision and recall are computed. Finally, the GLEU score is determined as the minimum of this precision and recall.
We follow the same tokenization process as discussed earlier and
use the \textit{Huggingface} library\footnote{\url{https://github.com/huggingface/evaluate/tree/main/metrics/google_bleu}} to compute GLEU scores.

\noindent
$\bullet$ \textbf{chrF$++$}: chrF++ is another automatic MT evaluation metric based on character n-grams based precision \& recall enhanced with word n-grams. The F-score is computed by averaging (arithmetic mean) across all character and word n-grams, with the default order for character n-grams set at 6 and word n-grams at 2. It has been shown to have better correlation with human judgements~\citep{popovic2017chrf++}. Here also, we use the \textit{SacreBLEU} library (stated above) to compute chrF++ score.\footnote{\textit{SacreBLEU} signature for chrF++: $CHRF+nrefs:1+case:mixed+eff:yes+nc:6+nw:2+space:no+version:2.2.0$.}

We scale all the three metric values to the range $[0,100]$, to make them more comparable.

\begin{table*}[htb]
\centering
\captionsetup{width=\textwidth}
\caption{Corpus-level BLEU, GLEU, and chrF++ scores for all MT systems, over the three {\corpus} datasets for the languages Bengali (BN), Hindi (HI), Tamil (TA), and Marathi (MR). All values are averaged over all text-pairs in a particular dataset (except for Davinci-003 and GPT-3.5T-Inst; details in the text). For each dataset, and each English-Indian language pair, the best value of each metric is boldfaced.}
\begin{tabular}{|c|c|ccc|ccc|ccc|}
\hline
\multirow{2}{*}{EN $\rightarrow$ IN} & \multirow{2}{*}{Model} & \multicolumn{3}{c|}{{\IPdataset}} & \multicolumn{3}{c|}{{\Actsdataset}} & \multicolumn{3}{c|}{{\CCIdataset}} \\
& & BLEU & GLEU & chrF++ & BLEU & GLEU & chrF++ & BLEU & GLEU & chrF++\\
\hline
\multirow{9}{*}{EN $\rightarrow$ BN} & GOOG & 27.7 & 30.7 & 56.8 & 12.0 & 17.0 & 40.7 & \textbf{52.0} & \textbf{53.6} & \textbf{74.8}\\
& MSFT & \textbf{31.0} & \textbf{33.8} & \textbf{59.4} & 18.4 & \textbf{23.1} & \textbf{45.6} & 36.5 & 40.4 & 66.2 \\
& Davinci-003 & 12.4 & 17.1 & 40.7 & 7.0 & 11.8 & 31.5 & 12.4 & 17.5 & 39.0 \\
& GPT-3.5T-Inst & 14.5 & 18.3 & 43.2 & 8.1 & 12.6 & 33.2 & 14.9 & 19.1 & 41.9 \\
& NLLB-1.3B & 25.6 & 28.4 & 54.0 & 13.9 & 19.2 & 40.9 & 18.3 & 22.4 & 42.6 \\
& NLLB-3.3B & 23.7 & 27.5 & 53.6 & 13.8 & 19.0 & 40.9 & 18.6 & 22.3 & 42.2 \\
& IndicTrans & 24.7 & 27.3 & 51.7 & \textbf{18.6} & 21.8 & 45.5 & 20.9 & 25.6 & 50.2 \\
& mBART-50 & 2.1 & 4.1 & 21.1 & 0.6 & 2.7 & 19.6 & 2.7 & 3.9 & 18.6 \\
& OPUS & 4.5 & 9.8 & 28.2 & 4.2 & 9.3 & 25.7 & 3.0 & 7.0 & 20.4\\
\hline
\multirow{9}{*}{EN $\rightarrow$ HI} & GOOG & 36.6 & 35.3 & 53.8 & 21.2 & 26.7 & 47.1 & 46.0 & 48.4 & 67.3\\
& MSFT & \textbf{38.5} & \textbf{37.0} & \textbf{54.9} & \textbf{46.4} & \textbf{48.9} & \textbf{67.3} & 45.5 & 48.2 & \textbf{67.5} \\
& Davinci-003 & 21.1 & 23 & 42.9 & 15.9 & 20.7 & 39.9 & 25.1 & 28.8 & 47.3 \\
& GPT-3.5T-Inst & 26.3 & 27.6 & 46.6 & 16.9 & 21.0 & 42.7 & 31.6 & 34.1 & 54.0 \\
& NLLB-1.3B & 32.3 & 32.3 & 51.2 & 34.0 & 37.6 & 56.7 & 31.2 & 33.7 & 49.8 \\
& NLLB-3.3B & 34.3 & 33.9 & 52.4 & 36.1 & 39.5 & 58.6 & 33.8 & 35.4 & 51.5 \\
& IndicTrans & 27.0 & 28.1 & 45.1 & 45.7 & 48.2 & 66.6 & \textbf{49.1} & \textbf{49.8} & 67.1 \\
& mBART-50 & 23.9 & 26.1 & 45.5 & 43.6 & 46.0 & 64.8 & 31.0 & 34.0 & 52.0 \\
& OPUS & 9.8 & 14.2 & 28.3 & 9.3 & 16.4 & 30.2 & 4.9 & 10.2 & 21.1\\
\hline
\multirow{9}{*}{EN $\rightarrow$ TA} & GOOG & \textbf{39.3} & \textbf{41.8} & \textbf{69.4} & 8.1 & 13.7 & 37.0 & \textbf{41.4} & \textbf{44.0} & \textbf{70.7}\\
& MSFT & 35.3 & 38.7 & 68.8 & \textbf{12.1} & \textbf{17.6} & \textbf{46.3} & 29.5 & 33.7 & 64.9 \\
& Davinci-003 & 9.2 & 12.6 & 36.9 & 4.8 & 8.3 & 29.6 & 8.5 & 10.4 & 31.3 \\
& GPT-3.5T-Inst & 6.2 & 9.2 & 34.5 & 3.7 & 6.5 & 28.0 & 7.3 & 9.3 & 32.0 \\
& NLLB-1.3B & 34.2 & 36.7 & 66.5 & 8.7 & 14.9 & 40.6 & 17.0 & 21.0 & 45.2 \\
& NLLB-3.3B & 34.2 & 37.3 & 67.2 & 9.4 & 15.2 & 41.2 & 18.6 & 21.9 & 46.4 \\
& IndicTrans & 21.4 & 25.5 & 51.9 & 11.1 & 16.7 & 43.7 & 22.9 & 26.8 & 56.1 \\
& mBART-50 & 14.8 & 19.6 & 49.2 & 8.2 & 13.6 & 37.9 & 11.2 & 13.6 & 23.6 \\
& OPUS & 5.5 & 10.2 & 28.4 & 9.3 & 16.4 & 30.2 & 2.4 & 6.3 & 19.5\\
\hline
\multirow{9}{*}{EN $\rightarrow$ MR} & GOOG & \textbf{23.0} & \textbf{25.6} & \textbf{51.6} & 8.6 & 14.6 & 37.5 & \textbf{51.3} & \textbf{53.0} & \textbf{74.8}\\
& MSFT & 19.4 & 22.8 & 49.6 & \textbf{13.9} & \textbf{19.6} & \textbf{45.0} & 34.1 & 38.3 & 65.8 \\
& Davinci-003 & 7.6 & 11.4 & 34 & 4.5 & 7.9 & 29.1 & 11.7 & 15.0 & 35.2 \\
& GPT-3.5T-Inst & 9.5 & 13.6 & 37.3 & 3.5 & 6.1 & 26.5 & 9.3 & 12.0 & 33.7 \\
& NLLB-1.3B & 17.6 & 21.7 & 48.0 & 12.8 & 17.8 & 42.3 & 17.9 & 21.1 & 42.3 \\
& NLLB-3.3B & 18.6 & 21.5 & 47.9 & 12.8 & 18.2 & 42.6 & 20.2 & 23.5 & 44.8 \\
& IndicTrans & 16.0 & 19.6 & 44.0 & 12.9 & 18.5 & 42.1 & 28.2 & 32.0 & 56.7 \\
& mBART-50 & 1.9 & 4.1 & 23.6 & 1.7 & 4.6 & 26.4 & 3.7 & 5.1 & 23.6 \\
& OPUS & 4.0 & 8.2 & 24.7 & 3.3 & 8.4 & 23.1 & 2.8 & 6.2 & 18.0\\
\hline
\end{tabular}
\label{tab:MILPaC_MT_all_Scores_Part_1}
\end{table*}



\begin{table*}[htb]
\centering
\footnotesize
\captionsetup{width=\textwidth}
\caption{Corpus-level BLEU, GLEU, and chrF++ scores for all MT systems, over \IPdataset{} and \Actsdataset{} datasets for the languages Telugu (TE), Malayalam (ML), Punjabi (PA), Oriya (OR), and Gujarati (GU). All values are averaged over all text-pairs in a particular dataset. 
For each dataset, and each English-Indian language pair, the best value of each metric is boldfaced.}
\begin{tabular}{|c|c|ccc|ccc|}
\hline
\multirow{2}{*}{EN $\rightarrow$ IN} & \multirow{2}{*}{Model} & \multicolumn{3}{c|}{{\IPdataset}} & \multicolumn{3}{c|}{{\Actsdataset}} \\
& & BLEU & GLEU & chrF++ & BLEU & GLEU & chrF++ \\
\hline
\multirow{9}{*}{EN $\rightarrow$ TE} & GOOG & \textbf{22.4} & \textbf{23.2} & \textbf{48.9} & 6.6 & 11.4 & 28.8 \\
& MSFT & 15.8 & 18.3 & 44.8 & \textbf{12.0} & \textbf{16.9} & 39.4 \\
& Davinci-003 & 6.2 & 9.7 & 29.7 & 4.2 & 7.2 & 22.7 \\
& GPT-3.5T-Inst & 4.4 & 6.8 & 27.9 & 3.5 & 6.0 & 23.0 \\
& NLLB-1.3B & 18.2 & 20.2 & 45.8 & 8.0 & 13.4 & 33.5 \\
& NLLB-3.3B & 18.7 & 20.7 & 46.3 & 7.5 & 12.8 & 33.8\\
& IndicTrans & 15.5 & 17.6 & 40.6 & 11.9 & 16.8 & \textbf{40.4} \\
& mBART-50 & 1.4 & 3.8 & 10.9 & 3.4 & 7.6 & 15.7 \\
& OPUS & 3.7 & 7.4 & 22.7 & 5.4 & 9.9 & 23.7 \\
\hline
\multirow{9}{*}{EN $\rightarrow$ ML} & GOOG & 22.3 & 27.7 & 57.5 & 7.3 & 12.4 & 32.2 \\
& MSFT & \textbf{34.2} & \textbf{37.7} & \textbf{66.5} & 10.8 & 17.0 & 46.2 \\
& Davinci-003 & 5.1 & 8.5 & 28.9 & 4.0 & 7.0 & 24.8 \\
& GPT-3.5T-Inst & 5.3 & 8.7 & 31.6 & 2.5 & 4.3 & 21.2 \\
& NLLB-1.3B & 25.6 & 29.7 & 60.0 & 8.0 & 13.0 & 39.3 \\
& NLLB-3.3B & 25.3 & 29.6 & 60.5 & 7.5 & 12.7 & 38.7 \\
& IndicTrans & 19.8 & 24.5 & 48.9 & \textbf{16.6} & \textbf{21.2} & \textbf{50.3} \\
& mBART-50 & 2.6 & 5.7 & 20.3 & 4.4 & 8.7 & 23.9 \\
& OPUS & 3.9 & 8.1 & 25.4 & 5.4 & 10.1 & 24.8 \\
\hline
\multirow{9}{*}{EN $\rightarrow$ PA} & GOOG & 17.8 & 20.8 & 41.3 & 8.9 & 14.1 & 28.6 \\
& MSFT & \textbf{30.2} & \textbf{30.5} & \textbf{51.3} & \textbf{40.1} & \textbf{42.4} & \textbf{62.5} \\
& Davinci-003 & 9.1 & 12.9 & 31 & 6.4 & 10.7 & 26.8 \\
& GPT-3.5T-Inst & 13.4 & 16.2 & 35.5 & 6.9 & 10.7 & 29.1 \\
& NLLB-1.3B & 27.5 & 28.1 & 48.5 & 19.6 & 25.0 & 44.0 \\
& NLLB-3.3B & 29.6 & 29.8 & 49.8 & 20.5 & 25.9 & 45.0 \\
& IndicTrans & 28.1 & 28.8 & 47.6 & 24.0 & 28.8 & 48.8\\
& OPUS & 5.5 & 10.3 & 22.3 & 6.7 & 12.6 & 25.7 \\
\hline
\multirow{9}{*}{EN $\rightarrow$ OR} & GOOG & 2.4 & 6.5 & 29.0 & 4.1 & 8.2 & 26.3 \\
& MSFT & 5.5 & 9.0 & 33.7 & 7.6 & 13.3 & 37.3 \\
& Davinci-003 & 3.3 & 6.8 & 25.8 & 3.3 & 6.1 & 22.5 \\
& GPT-3.5T-Inst & 2.1 & 4.4 & 22.4 & 1.9 & 4.3 & 20.8 \\
& NLLB-1.3B & 5.3 & 8.8 & 33.1 & 9.5 & 14.2 & 37.0 \\
& NLLB-3.3B & \textbf{6.3} & \textbf{10.1} & \textbf{34.6} & \textbf{10.1} & \textbf{15.5} & 39.3 \\
& IndicTrans & 4.9 & 8.6 & 30.5 & 8.9 & 15.0 & \textbf{40.4} \\
& OPUS & 2.3 & 5.2 & 21.0 & 3.4 & 7.5 & 22.9 \\
\hline
\multirow{9}{*}{EN $\rightarrow$ GU} & GOOG & 43.6 & 46.0 & 67.8 & 14.3 & 19.5 & 42.1 \\
& MSFT & \textbf{47.3} & \textbf{49.2} & \textbf{70.6} & 21.7 & 26.1 & \textbf{51.9} \\
& Davinci-003 & 12.3 & 17.3 & 37.2 & 8.5 & 12.3 & 32.0 \\
& GPT-3.5T-Inst & 15.8 & 19.5 & 40.9 & 5.4 & 10.1 & 31.5 \\
& NLLB-1.3B & 43.5 & 45.6 & 66.4 & 20.3 & 24.4 & 49.1 \\
& NLLB-3.3B & 42.6 & 44.8 & 66.2 & 19.7 & 24.2 & 49.8\\
& IndicTrans & 31.3 & 34.9 & 56.3 & \textbf{22.9} & \textbf{27.0} & 50.9\\
& mBART-50 & 1.7 & 3.2 & 5.2 & 0.6 & 3.5 & 3.0 \\
& OPUS & 9.2 & 14.7 & 30.0 & 5.1 & 9.8 & 23.7 \\
\hline
\end{tabular}
\label{tab:MILPaC_MT_all_Scores_Part_2}
\end{table*}


\subsection{Results of automatic evaluation}

Table~\ref{tab:MILPaC_MT_all_Scores_Part_1} shows the performances of all the MT systems across the 3 datasets of \corpus{} for the languages  Bengali (BN), Hindi (HI), Tamil (TA), and Marathi (MR). Table~\ref{tab:MILPaC_MT_all_Scores_Part_2} shows the performances of all the MT systems over \IPdataset{} and \Actsdataset{} for the languages Telugu (TE), Malayalam (ML), Punjabi (PA), Oriya (OR), and Gujarati (GU) (since \CCIdataset{} does not have these languages).\footnote{All experiments were conducted on a machine having 2 x NVIDIA Tesla P100 16GB GPU.}


Note that the Davinci-003 and GPT-3.5T-Inst have a maximum bound on the number of tokens (as stated earlier).
Hence, out of the 1,460 EN-IN pairs in the \CCIdataset{} dataset,  the 77 longest text units could not be fed into the Davinci-003 and GPT-3.5T-Inst models in their entirety (these include 19 text units in BN, 15 in HI, 31 in TA, and 12 in MR). 
One option was to break these long text units into chunks and then translate chunk-wise, but we preferred not to translate chunk-wise in order to get a fair comparison with other models. 
Hence, the 77 longest text units were not considered while evaluating Davinci-003 and GPT-3.5T-Inst models over the \CCIdataset{} dataset.
For all other MT systems, the values reported are averaged over all text-pairs in each dataset.

We find that no single model performs the best in all scenarios.
MSFT, GOOG, and IndicTrans are the 3 best models that generally perform the best in most scenarios. 
It can be noted that the same trend was observed by \citep{ramesh2021samanantar} for Indian non-legal text, i.e., IndicTrans, GOOG, and MSFT outperform mBART-50 and OPUS.
A notable exception is that NLLB performs the best for EN-OR translation, possibly because Oriya is a resource-poor language that NLLB is specially designed to handle. 

The translation performances of both Davinci-003 and GPT-3.5T-Inst are relatively poor. One particular problem we observed in the translations using these two LLMs is that, for longer text units, both models sometimes repeat multiple times the translation of the last few words in their translated text.

The scores for \Actsdataset{} are consistently lower than those for other datasets. This is expected since \Actsdataset{} has very formal legal language, which is challenging for all MT systems. Interestingly, though MSFT and GOOG perform the best over most datasets, IndicTrans performs better over \Actsdataset{} for several Indian languages (e.g., Malayalam \& Gujarati). 
The superior performance of IndicTrans over \Actsdataset{} may stem partly from the fact that it was trained on some legal documents from Indian government websites (such as State Assembly discussions) according to~\cite{ramesh2021samanantar}. 
However, it is {\it not} known publicly over what data commercial systems such as GOOG and MSFT are trained.

\subsection{Evaluation through Law practitioners} \label{sub:expert-evaluation}


Apart from the automatic evaluation of translation performance (as detailed above), we also evaluate the performance of MT systems through a set of Law students. This is an important part of the evaluation, since we can get an idea of how well automatic translation performance metrics agree with the opinion of domain experts. 

\noindent
\textbf{Recruiting Law students:} We recruited senior LLB \& LLM students from the Rajiv Gandhi School of Intellectual Property Law which is one of the most reputed Law schools in India. 
We recruited these students based on the recommendations of a Professor of the same Law school, and considering two requirements --
(i)~the students should have good Legal domain knowledge; (ii)~they should be fluent in English and be \textit{native speakers of some Indian language} (for which they will evaluate translations). 
We could obtain Law students who are native speakers of only four Indian languages. 
Specifically, we recruited 5 Law students in total, among whom 2 are native speakers of Hindi, and the other three are native speakers of Bengali, Tamil, and Marathi (1 for each language).
Hence, this human evaluation is carried out for these 4 Indian languages only. 

\noindent
\textbf{Survey Setup:} Given the limited availability of Law students, only the 3 MT systems, GOOG, MSFT, and IndicTrans (the best performers in our automatic evaluation), were evaluated.
\CCIdataset{} and \Actsdataset{} were selected for this human evaluation since these are the more challenging datasets.
We randomly selected 50 English text units from each of \CCIdataset{} and \Actsdataset{} datasets, and their translations in Hindi, Bengali, Tamil, and Marathi by the 3 MT systems stated above.
Then, a particular Law student was shown these 50 randomly selected English text units (from each of \CCIdataset{} and \Actsdataset{}) datasets, and their translations (by the said systems) in his/her native language, and asked to evaluate the translation quality.

The Law students who participated in our human surveys were clearly informed of the purpose for which the annotations/surveys were being carried out. They consented to participation in the survey and agreed that automatic translation tools can be of great help in the Indian legal scenario.
Each of them was paid a mutually agreed honorarium for the evaluation. 
We took all possible steps to ensure that their annotations are unbiased (e.g., by anonymizing the names of the MT systems whose outputs they were annotating) and accurate (e.g., by asking them to write justifications for low scores, post-survey discussions, etc.).
We also conducted a pilot survey (described below) and discussed with them after their evaluation to get their feedback. 
Through all these steps, we attempted to ensure that the annotations are done rigorously.

The Law student who is a native speaker of Bengali, was not available during the evaluation of \Actsdataset{}. 
Thus, in total, we performed human evaluations for 1,050 text pairs [(50 text units of \CCIdataset{} x 3 MT systems x 4 language pairs) + (50 text units of \Actsdataset{} x 3 MT systems x 3 language pairs)].

\noindent
\textbf{Metrics for human evaluation:} 
Typically, manual evaluation of MT systems use metrics such as Accuracy / Adequacy 
and Fluency~\citep{escribe-2019-human}. 
Since we are considering the domain of Law, we asked the Law students how they would like to evaluate translations.
To this end, an initial pilot survey had been conducted;
through this pilot survey and subsequent discussion, we agreed upon the following 3 metrics for the evaluation of legal translations: \\
(1)~\underline{\textit{Preservation of Meaning (POM)}}: similar to the standard measure of Accuracy/Adequacy, this captures how well the translation captures all of the meaning/information in the source text. \\
(2)~\underline{\textit{Suitability for Legal Use (SLU)}}: a domain-specific metric, that checks if the translation is sufficiently formal and specific/unambiguous to be used in Legal drafting. Note that SLU has some differences with POM, which is explained later in this section. \\
(3)~\underline{\textit{Fluency (FLY)}}: how grammatically correct and fluent the translations are.\\ 
All three metrics were recorded on a Likert scale of 1-5, with 1 being the lowest score (worst translation performance) and 5 being the highest (best translation performance). 

\begin{table}[tb]
\captionsetup{width=\textwidth}
\caption{Mean scores given by Law students (in [1--5]), and the automatic metrics (in [0--100]) for the same EN-IN pairs from  \CCIdataset{} and \Actsdataset{}.
Each value is averaged over 50 randomly selected text units from a dataset. The best values are boldfaced.}
\begin{tabular}{|c|c|ccc|ccc|}
\hline
\multirow{2}{*}{EN $\rightarrow$ IN} & \multirow{2}{*}{Model} & \multicolumn{3}{c|}{Automatic Metrics} & \multicolumn{3}{c|}{Human Scores} \\
& & BLEU & GLEU & chrF++ & POM & SLU & FLY \\
\hline
\hline
\multicolumn{8}{|c|}{\CCIdataset{}} \\
\hline
\multirow{3}{*}{EN $\rightarrow$ BN} & IndicTrans & 19.4 & 24.3 & 46.5 & 3.50 & 3.48 & 3.52 \\
& MSFT & 33.7 & 38.6 & 64.3 & \textbf{3.56} & \textbf{3.54} & \textbf{3.56} \\
& GOOG & \textbf{50.6} & \textbf{52.4} & \textbf{73.5} & 3.48 & 3.40 & 3.46 \\
\hline
\multirow{3}{*}{EN $\rightarrow$ HI} & IndicTrans & \textbf{49.4} & \textbf{50.6} & \textbf{66.2} & \textbf{3.67} & \textbf{3.53} & \textbf{3.97}\\
& MSFT & 44.9 & 47.9 & 66.8 & 2.20 & 1.95 & 2.23 \\
& GOOG & 44.7 & 47.7 & 66.1 & 2.19 & 2.16 & 2.13 \\
\hline
\multirow{3}{*}{EN $\rightarrow$ MR} & IndicTrans & 28.9 & 32.6 & 56.1 & 4.02 & 4.02 & 4.02\\
  & MSFT & 32.6 & 37.1 & 64.6 & \textbf{4.06} & \textbf{4.06} & \textbf{4.04} \\
  & GOOG & \textbf{49.9} & \textbf{52.0} & \textbf{73.7} & 3.54 & 3.46 & 3.52 \\
\hline
\multirow{3}{*}{EN $\rightarrow$ TA} & IndicTrans & 22.0 & 26.4 & 54.6 & \textbf{2.98} & \textbf{2.86} & \textbf{2.98} \\
& MSFT & 27.0 & 31.5 & 62.0 & 2.86 & 2.74 & 2.86 \\
& GOOG & \textbf{38.8} & \textbf{41.5} & \textbf{67.8} & 2.80 & 2.80 & 2.80 \\
\hline 
\multicolumn{8}{|c|}{\Actsdataset{}} \\
\hline
\multirow{3}{*}{EN $\rightarrow$ HI} & IndicTrans & 47.1 & 49.5 & \textbf{69.3} & \textbf{4.65} & \textbf{4.07} & \textbf{4.74}\\
& MSFT & \textbf{47.4} & \textbf{49.6} & 68.7 & 4.32 & 3.80 & 4.54 \\
& GOOG & 18.4 & 24.2 & 44.7 & 3.13 & 2.52 & 3.25 \\
\hline
\multirow{3}{*}{EN $\rightarrow$ MR} & IndicTrans & \textbf{18.7} & \textbf{23.7} & \textbf{47.7} & \textbf{4.0} & \textbf{3.46} & \textbf{3.90}\\
& MSFT & 15.7 & 21.6 & 47.1 & 3.60 & 2.46 & 3.56 \\
& GOOG & 10.4 & 16.5 & 39.9 & 1.94 & 1.16 & 1.9 \\
\hline
\multirow{3}{*}{EN $\rightarrow$ TA} & IndicTrans & 12.3 & 17.1 & 43.0 & \textbf{4.0} & \textbf{3.76} & \textbf{4.02}\\
& MSFT & \textbf{13.1} & \textbf{18.0} & \textbf{43.3} & 3.84 & 3.72 & 3.88 \\
& GOOG & 10.1 & 15.0 & 36.0 & 3.02 & 2.78 & 3.06 \\
\hline
\end{tabular}
\label{tab:Expert-Scores-CCI-Acts}
\end{table}

\noindent
\textbf{Results of the human survey:}
Table~\ref{tab:Expert-Scores-CCI-Acts} shows the mean scores (POM, SLU, FLY) given by the Law students to each MT system for \CCIdataset{} \& \Actsdataset{}.
For comparison, we also report the automatic metrics averaged over the same text-pairs that the Law students evaluated. 
For \CCIdataset{}, the Law students preferred IndicTrans for Hindi \& Tamil, and MSFT for Bengali \& Marathi, even though GOOG performs better for most of these languages according to automatic metrics. 
For \Actsdataset{}, the Law students preferred the IndicTrans translations for all three Indian languages, even though MSFT obtained the highest BLEU \& GLEU scores for EN-TA and EN-HI translations.
The good performance of IndicTrans over \Actsdataset{} may be because IndicTrans has been pretrained on legislative documents in some Indian languages~\citep{ramesh2021samanantar}.
However, performance of most MT systems over \CCIdataset{} is {\it not} of much satisfaction to the Law students (apart from EN-MR translation).


\noindent
\textbf{Comparing the two metrics for human survey (SLU vs POM):} For human evaluation of Machine Translation in general domains, the two metrics Accuracy \& Fluency are deemed to be sufficient. Accuracy checks whether the meaning has been rendered correctly in the translation, while Fluency checks whether the translation is grammatically correct.

However, we understood from our discussions with the Law students that, for evaluating translations in the Legal domain, a third metric (SLU: Suitability for Legal Use) is also essential. 
A translation may perfectly capture the meaning of the source text, but if the language of the translation is not keeping with the Legal standards, then the translated text cannot be used for purposes such as legal drafting. 
This difference arises mainly because legal language needs to be very formal and unambiguous, particularly while drafting laws and guidelines.
Sometimes exact words and phrases need to be used, whereas if a translation uses a synonym or a similar phrase, then the translation will not be considered suitable for legal use, even if the meaning of the source text is well captured. 
This is why, after discussion with the Law students, we decided to use two separate metrics -- \textit{Preservation of Meaning} (POM) and \textit{Suitability of Legal Use} (SLU) -- during our human survey.

It can be seen from Table~\ref{tab:Expert-Scores-CCI-Acts} that for the \Actsdataset{} dataset, SLU scores are substantially lower than the POM scores, for all Indian languages and all MT systems. 
This means that even when the translations well captured the meanings of the source text, the translated texts are not sufficiently formal for legal drafting, as per the opinion of the Law students. 
This problem is particularly important for the \Actsdataset{} dataset which contains very formal legal text that the MT systems often fail to generate.
On the other hand, the SLU scores are much closer to the POM scores for the \CCIdataset{} dataset, possibly because this dataset contains simplified versions of legal text.
It is an interesting direction of future research as to how the legal translations can be made sufficiently formal when necessary, e.g., when the formal laws of a country are to be translated.

\noindent
\textbf{Inter Annotator Agreement (IAA):} We could compute IAA only for English-to-Hindi translations since we had 2 annotators only for Hindi.
For each dataset (\CCIdataset{} and \Actsdataset{}), we considered all the 150 (50 text-pairs $\times$ 3 systems) EN-HI translations for which 2 annotators gave ratings, and computed the \textit{Pearson's correlation} between the sets of scores given by the two annotators.
The IAA values show relatively high agreement for \CCIdataset{} (agreement for POM: 0.68, SLU: 0.71, FLY: 0.70; average IAA: 0.70) and moderate agreement for \Actsdataset{} (agreement for POM: 0.52, SLU: 0.43, FLY: 0.55; average IAA: 0.50).

It is to be noted that different Law practitioners have different preferences about translations of legal text (as shown by the moderate IAA between the two annotators for English-to-Hindi translations). Hence the observations presented above are according to the Law students we consulted, and can vary in case of other Law practitioners.

\begin{table}[tb]
\centering
\captionsetup{width=\textwidth}
\caption{Pearson's correlation values of automatic metrics with human-assigned scores for \CCIdataset{} and \Actsdataset{}. Most correlations are low, apart from those for English-to-Hindi translations.}
\small
\begin{tabular}{|c|c|cc|}
\hline
\multirow{2}{*}{EN $\rightarrow$ IN} & \multirow{2}{*}{Human Metrics} & \multicolumn{2}{c|}{Automatic Metrics} \\
& & GLEU & chrF++ \\
\hline
\hline
\multicolumn{4}{|c|}{\CCIdataset{}} \\
\hline
\multirow{3}{*}{EN $\rightarrow$ BN} & POM & 0.117 & 0.138 \\
& SLU & 0.105 & 0.130 \\
& FLY & 0.105 & 0.122 \\
\hline
\multirow{3}{*}{EN $\rightarrow$ HI} & POM & 0.262 & 0.208 \\
& SLU & 0.238 & 0.188 \\
& FLY & 0.222 & 0.170 \\
\hline
\multirow{3}{*}{EN $\rightarrow$ MR} & POM & $-0.178$ & $-0.183$ \\
 & SLU & $-0.149$ & $-0.179$ \\
 & FLY & $-0.162$ & $-0.175$ \\
\hline
\multirow{3}{*}{EN $\rightarrow$ TA} & POM & 0.180 & 0.186 \\
& SLU & 0.208 & 0.208 \\
& FLY & 0.180 & 0.186 \\
\hline
\multirow{3}{*}{Overall} & POM & 0.065 & 0.097 \\
& SLU & 0.057 & 0.091 \\
& FLY & 0.070 & 0.092 \\
\hline
\multicolumn{4}{|c|}{\Actsdataset{}} \\
\hline
\multirow{3}{*}{EN $\rightarrow$ HI} & POM & 0.602 & 0.648 \\
& SLU & 0.648 & 0.687 \\
& FLY & 0.607 & 0.650 \\
\hline
\multirow{3}{*}{EN $\rightarrow$ MR} & POM & 0.203 & 0.272 \\
& SLU & 0.269 & 0.280 \\
& FLY & 0.194 & 0.269 \\
\hline
\multirow{3}{*}{EN $\rightarrow$ TA} & POM & 0.164 & 0.314 \\
& SLU & 0.135 & 0.282 \\
& FLY & 0.147 & 0.314 \\
\hline
\multirow{3}{*}{Overall} & POM & 0.378 & 0.435 \\
& SLU & 0.335 & 0.377 \\
& FLY & 0.403 & 0.457 \\
\hline
\end{tabular}
\label{tab:correlation_table_CCI_Acts}
\end{table}

\noindent
\textbf{Correlation of automatic metrics with human scores:}
Table~\ref{tab:Expert-Scores-CCI-Acts} brought out some differences in the MT system performances according to the automatic metrics and the scores given by the Law students (as stated above). 
These differences led us to investigate the correlation between the automatic metrics (GLEU and chrF++) and human scores (POM, SLU, FLY).
Corresponding to each Indian language and each dataset, we checked the Pearson's correlation over the $150$ data points that were manually evaluated (50 text-pairs $\times$ 3 MT systems). 

The correlation values corresponding to each language pair are given in Table~\ref{tab:correlation_table_CCI_Acts} for \CCIdataset{} and \Actsdataset{} datasets respectively.\footnote{As stated earlier, the Law student who is a native speaker of Bengali, was not available during the expert evaluation of \Actsdataset{}.} We observe the correlations of automatic metrics and human scores to be moderate for \Actsdataset{} (between $0.35$ and $0.4$ on average). Only for EN-HI, correlations are higher (above $0.6$). 
Whereas, for \CCIdataset{}, all correlations were very low, around $0.1$ on average. 
Among the automatic metrics, chrF++ shows slightly higher correlation with human scores than GLEU. 
Such low correlations with human scores indicate the need for improvement in the automatic evaluation of MT systems in the legal domain. 


\subsection{Analysis of errors committed by the MT systems}\label{subsec:MT_errors}

As part of our survey (that has been described earlier), we also asked the Law students to give, along with the POM, SLU, and FLY scores, comments to explain why they gave certain translations low scores. We also asked them to identify different types of errors committed by the MT systems. 
They identified various types of errors such as \textit{inclusion of extra words in translations}, \textit{mis-translation of terms}, \textit{presence of untranslated portions}, \textit{consistency errors} (inconsistency between the factual information in the original and translated text), and so on.
Table~\ref{tab:MT-error-analysis} shows the common errors and the percentage of text-pairs in which each error was identified by the Law students (across text-pairs in all Indian languages, that were evaluated by them). 
GOOG and MSFT frequently have untranslated portions (which possibly led to their poor expert scores).
Most errors by IndicTrans were minor punctuation errors, which did not affect the translation quality much; however IndicTrans often mis-translates legal terms. 
We hope that these observations will help in improving these MT systems in future. 

\begin{table}[tb]
    \centering
    \captionsetup{width=\textwidth}
    \caption{Occurrence of errors in translations committed by the 3 MT systems GOOG, MSFT and IndicTrans, as detected by the Law students. Relatively high values are highlighted in boldface.} 
    \begin{tabular}{|l|ccc|}
         \hline
         Error & GOOG & MSFT & IndicTrans \\
         \hline
         Extra words included &  0.7\% & 0.4\% & 0.9\%\\
         Omission of words &  0.9\% & 0.0\% & 0.9\% \\
         General terms mistranslated & 0.9 \% & 1.3\% & 1.1\%\\
         Legal terms mistranslated & 1.6\% & 3.8\% & \textbf{5.1\%} \\
         Untranslated Portions & \textbf{10.9\%} & \textbf{7.3\%} & 2.7\% \\
         Grammatical Errors & 1.1\% & 0.2\% & 0.0\%\\
         Spelling Errors & 0.2\% & 0.2\% & 0\% \\
         Punctuation Errors & 0.4\% & 1.6\% & \textbf{6.9\%} \\
         Consistency Errors & 2.4\% & 0.7\% & 0.9\%\\
         \hline
    \end{tabular}
    \label{tab:MT-error-analysis}
\end{table}

\begin{figure*}[tb]
\centering
\captionsetup{width=\textwidth}
\includegraphics[width=\textwidth]{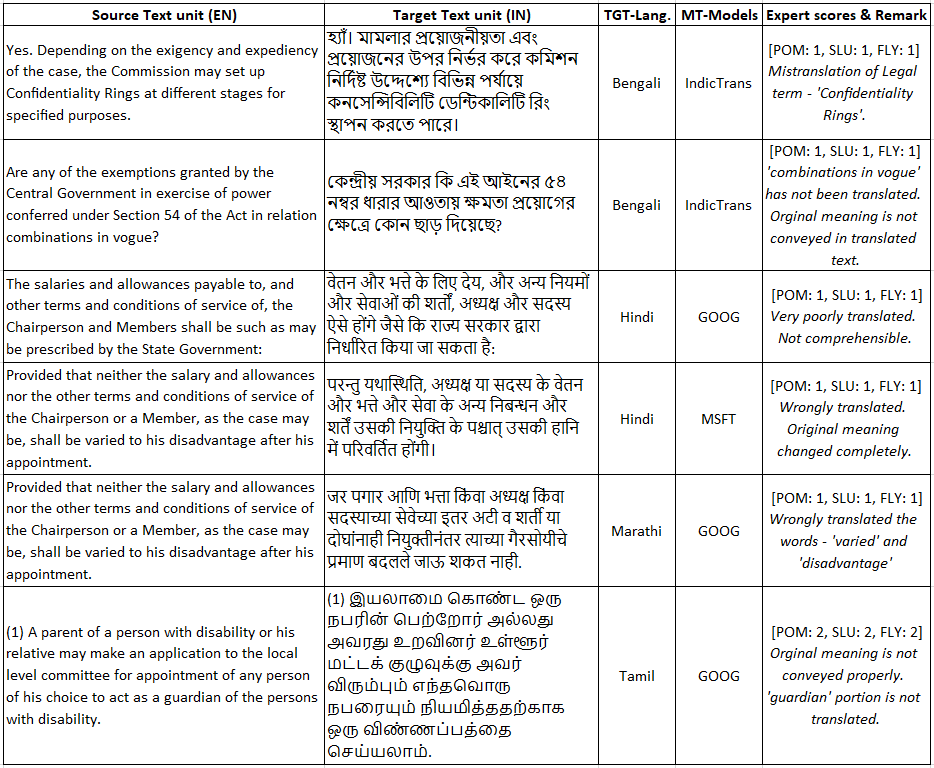}
\caption{Examples of errors committed by various MT systems across the \corpus{} datasets. The examples are selected from among the textual units that were used in the surveys by the Law students and obtained low scores. 
The $3^{rd}$ column states the target language (in which the translated text unit in the $2^{nd}$ column is), and the $4^{th}$ column states the MT system. 
The last column gives the expert scores in the range $[0, 5]$ (the metrics POM, SLU, FLY are explained in the text) and a remark/justification given by the expert.}
\label{fig:Error_MT_systems}
\end{figure*}

Some examples of errors committed by the MT systems have been shown in Figure~\ref{fig:Error_MT_systems}. 
For each example, the POM, SLU, and FLY scores, along with the comments written by the Law students (who gave the scores), explaining the problem(s) / error(s) in the translated text, are reported.
We see examples where the Law students opined that the meaning of the original text has not been captured in the translation, examples where a portion of the original text has not been translated, and examples where certain terms having domain-specific significance have not been properly translated.

\section{Other applications of \corpus{}}
Till now, we have used \corpus{} to benchmark MT systems for English-to-Indian languages translation of legal text. In addition to this, the \corpus{} corpus has several other potential applications, some of which we describe briefly in this section.

\begin{table}[tb]
\centering
\captionsetup{width=\textwidth}
\caption{Evaluating Indian-to-Indian translation using {\IPdataset}. The best values are in boldface.}
\begin{tabular}{|c|c|c|ccc|}
\hline
IN $\rightarrow$ IN & MODEL & MODE & BLEU & GLEU & chrF++ \\
\hline
\multirow{4}{*}{PA $\rightarrow$ MR} & \multirow{2}{*}{MSFT} & PA $\rightarrow$ MR & 17.2 & 20.2 & 44.5 \\
& & PA $\rightarrow$ EN $\rightarrow$ MR & \textbf{17.2} & \textbf{20.3} & \textbf{44.6} \\
\cline{2-6}
& \multirow{2}{*}{GOOG} & PA $\rightarrow$ MR    & \textbf{15.1} & \textbf{18.6} & \textbf{42} \\
& & PA $\rightarrow$ EN $\rightarrow$ MR & 13.3 & 16.8 & 39.5 \\
\hline
\multirow{4}{*}{GU $\rightarrow$ MR} & \multirow{2}{*}{MSFT} & GU $\rightarrow$ MR & 15.8 & 20.4 & 46.5 \\
& & GU $\rightarrow$ EN $\rightarrow$ MR & \textbf{16.0} & \textbf{20.5} & \textbf{46.6} \\
\cline{2-6}
& \multirow{2}{*}{GOOG} & GU $\rightarrow$ MR & \textbf{19.6} & \textbf{23.6} & \textbf{49.1} \\
& & GU $\rightarrow$ EN $\rightarrow$ MR & 15.7 & 20.4 & 45.4 \\
\hline
\multirow{4}{*}{BN $\rightarrow$ TE} & \multirow{2}{*}{MSFT} & BN $\rightarrow$ TE & 20.3 & 22.9 & 49.4 \\
& & BN $\rightarrow$ EN $\rightarrow$ TE & \textbf{20.4} & \textbf{23.0} & \textbf{49.5} \\
\cline{2-6}
& \multirow{2}{*}{GOOG} & BN $\rightarrow$ TE    & \textbf{19.3} & \textbf{22.3} & \textbf{47.3} \\
& & BN $\rightarrow$ EN $\rightarrow$ TE & 16.0 & 20.1 & 43.8 \\
\hline
\end{tabular}
\label{tab:Indic-Indic-Evaluation}
\end{table}

\subsection{Evaluating Indian-to-Indian legal translations} \label{secn:Indic-Indic}

Translation of legal text from one Indian language to another may be necessary, e.g., in scenarios where a person who is a native of one Indian state (and understands the local language of his/her native state) becomes a party to a legal case in a lower court of another state having a different local language. 
\corpus{} can be used for evaluating Indian-to-Indian language translation in the legal domain as well. 
To demonstrate this, we investigate the question -- {\it while translating between two low-resource Indian languages, is it better to translate directly, or through a high-resource intermediary language?}.
While answering this question systematically will require a full-fledged study, here we briefly investigate the question
using only the {\IPdataset} dataset and the GOOG and MSFT systems.

We select a low-resource Indian language-pair (e.g., Punjabi to Marathi translation), and then consider two approaches for this translation --
(1)~direct translation (PA $\rightarrow$ MR) and (2)~translation through English as an intermediary language (PA $\rightarrow$ EN $\rightarrow$ MR). 
Table~\ref{tab:Indic-Indic-Evaluation} compares these two approaches for three Indian language-pairs -- Punjabi to Marathi translation, Gujarati to Marathi translation, and Bengali to Telugu translation. 
Interestingly, for GOOG, direct translation is consistently better, but for MSFT, going through English marginally improves performance. 
While more experimentation is needed for a detailed understanding, we here demonstrate that \corpus{} can be used to address such questions on legal translation from one Indian language to another.

\subsection{Use of \corpus{} in fine-tuning MT systems} \label{subsec:MILPaC_fine}
The datasets in \corpus{} are much smaller in size compared to general-domain parallel corpora, due to two primary reasons. 
First, developing such corpora requires the involvement of Law practitioners which is expensive and limited. 
Second, creating these datasets required a lot of manual effort, due to the poor quality of the documents (specially in the \Actsdataset{} dataset) and the consequent challenges in OCR of these documents (detailed in the Appendix). 
Thus, the \corpus{} datasets are not suitable for training Neural MT systems from scratch. 
Hence we are proposing this corpus mainly for the evaluation of MT systems.
However, in this section, we show that \corpus{} is also suitable for \textit{fine-tuning MT systems} to improve their performance over Indian Legal text.

\begin{table}[tb]
\centering
\captionsetup{width=\textwidth}
\caption{Performance of the off-the-shelf IndicTrans and the fine-tuned IndicTrans (IndicTrans-FT) over 20\% test split of \corpus{}. 
All values are averaged over all text pairs in the 20\% test split (language-wise). For each English-Indian language pair, the best value of each metric is boldfaced. The fine-tuned version achieves much better translation, thus showing the utility of \corpus{} for fine-tuning MT systems to improve their performance in the legal domain.}
\begin{tabular}{|c|ccc|ccc|}
\hline
\multirow{2}{*}{EN $\rightarrow$ IN} & \multicolumn{3}{c|}{IndicTrans} & \multicolumn{3}{c|}{IndicTrans-FT} \\ 
& BLEU & GLEU & chrF++ & BLEU & GLEU & chrF++ \\
\hline
\multirow{1}{*}{EN $\rightarrow$ BN} & 17.7 & 22.6 & 46.9 & \textbf{41.8} & \textbf{43.7} & \textbf{67.1} \\
\hline
\multirow{1}{*}{EN $\rightarrow$ HI} & 42.7 & 45.0 & 62.7 & \textbf{53.4} & \textbf{54.2} & \textbf{71.6} \\
\hline
\multirow{1}{*}{EN $\rightarrow$ MR} & 20.6 & 25.0 & 50.0 & \textbf{40.1} & \textbf{41.5} & \textbf{65.4} \\
\hline
\multirow{1}{*}{EN $\rightarrow$ TA} & 19.4 & 23.4 & 51.5 & \textbf{32.7} & \textbf{35.1} & \textbf{64.6} \\
\hline 
\end{tabular}
\label{tab:fin_MILPac}
\end{table}

To this end, we randomly split \corpus{} (combining all 3 datasets) in a 70 (train) - 10 (validation) - 20 (test) ratio language-wise. 
For instance, we combined the English-Bengali text-pairs from all 3 datasets, and then split this combined set to get the train-test-validation splits for the Bengali language.
Then, for each Indian language, we \textit{fine-tuned the open-source MT model IndicTrans}~\citep{ramesh2021samanantar} with the 70\% train set (combined 70\% of all 3 \corpus{} datasets).
We used the 10\% validation split (combined 10\% of all 3 datasets) to select the best checkpoint among different iterations and to prevent overfitting by \textit{early stopping} with the \textit{patience} value as 3. 
Then we checked the performance of the fine-tuned IndicTrans (which we call `IndicTrans-FT') over the 20\% test set (20\% of the 3 datasets combined). 

Table~\ref{tab:fin_MILPac} compares the performance of the off-the-shelf IndicTrans~\citep{ramesh2021samanantar} and the fine-tuned IndicTrans-FT over the 20\% test set for 4 Indian languages.
For every language, we find that IndicTrans-FT gives substantial improvements over the off-the-shelf IndicTrans. Thus, though the intended primary use of \corpus{} is to evaluate MT systems, these results show that \corpus{} can also be used to fine-tune general domain MT systems to improve their translations in the legal domain.

%% file: conclu.tex
\section{Conclusion} \label{sec:conclu}

This work develops the first parallel corpus for Indian languages in the legal domain. We benchmark several MT systems using the corpus, including both automatic evaluation and a human survey by Law students.
This work yields important insights for the future:
(1)~It is necessary to improve existing MT systems for effective use in the legal domain. We indicate where to improve, by listing common errors committed by some of the MT systems (Table~\ref{tab:MT-error-analysis}). 
(2)~Automatic metrics correlate well with human scores only for EN-to-HI translation, but not for other Indian languages; hence better automatic metrics need to be devised for the legal domain, especially for quantifying the suitability of translated text for legal drafting (SLU).
We also demonstrate the applicability of the \corpus{} corpus for improving MT systems by fine-tuning them.
We believe that the corpus as well as the insights derived from this study will be important to improve the translation of legal text to Indian languages, a task that is critical for making justice accessible to millions of Indians. 


Our study highlights that existing state-of-the-art MT systems often commit various errors in legal translations. Therefore, we do \textit{not} advocate for the direct replacement of human translators (such as law practitioners who manually translate legal documents) with existing MT models. Rather, the study is performed keeping in mind that automatic MT systems can significantly speed up the litigation process, which is highly beneficial in countries with overburdened legal systems, such as India. Therefore, we wish to advocate a \textit{human-in-the-loop} approach where, as a first step, an existing MT system (e.g., the top performers over MILPaC) can be used to generate initial drafts, significantly expediting the translation process. Then, these drafts should be reviewed by a domain expert to correct errors and to make the translations more precise where necessary. This \textit{human-in-the-loop} approach will harness the speed of MT systems while maintaining the quality and reliability ensured by domain experts.

%% file: appendix.tex

\section{Additional details about the \corpus{} dataset}
\label{sec:appendix-dataset-details}






\subsection{Variation in the number of text units for different language-pairs in \IPdataset{}} \label{sub:appendix-ipdataset-variation}

For the \IPdataset{} dataset, ideally the version in every language should have 57 QA pairs (the number of QA pairs in the English version), hence ideally there should be 114 text units for each language-pair. 
However, the number of textual units in Table~\ref{tab:MILPaC-IP-CCI-Counts} (upper triangular part) show some variation for some of the language-pairs. These variations are explained as follows.

Some of the PDFs in the other languages have one or two QA pairs missing, which result in the differences in the numbers of textual units. For instance, 2 QA pairs are missing in the Bengali (BN) version, and 1 different QA pair is missing in the Telugu (TE) version. 
So all numbers in the BN row are at most $110~[114 - (2 \times 2)]$ due to the two missing QA pairs. Further, since the TE version also has another QA pair missing, the number of text units for BN-TE is $108~[114 - (3 \times 2)]$. 

Instead of giving so many details in the main text, we just mentioned ``approximately 57'' QA pairs in each version. We have manually mapped the existing QA pairs correctly for all language versions.

\subsection{List of Acts considered for the {\Actsdataset} dataset}
\label{sec:appendix-acts}

We used the following Indian judiciary Acts (statutes or written laws) to develop the \Actsdataset{} dataset. As stated in the main text, the selection of these Acts were based on their availability in multiple Indian languages, and the quality of the PDF scans available.
\begin{itemize}
    \small
    \item THE REGISTRATION OF BIRTHS AND DEATHS ACT, 1969
    \item THE BONDED LABOUR SYSTEM (ABOLITION) ACT, 1976
    \item THE SPICES BOARD ACT, 1986
    \item THE NATIONAL TRUST FOR WELFARE OF PERSONS WITH AUTISM, CEREBRAL PALSY, MENTAL RETARDATION AND MULTIPLE DISABILITIES ACT, 1999
    \item THE RIGHT TO INFORMATION ACT, 2005
    \item THE COMMISSIONS FOR PROTECTION OF CHILD RIGHTS ACT, 2005
    \item THE PRIVATE SECURITY AGENCIES (REGULATION) ACT, 2005
    \item THE SCHEDULED TRIBES AND OTHER TRADITIONAL FOREST DWELLERS (RECOGNITION OF FOREST RIGHTS) ACT, 2006
    \item THE MAINTENANCE AND WELFARE OF PARENTS AND SENIOR
    CITIZENS ACT, 2007
    \item THE NATIONAL INVESTIGATION AGENCY ACT, 2008
\end{itemize}
We downloaded the English and Indian language versions of these acts from \url{https://legislative.gov.in/central-acts-in-regional-language/}.

\section{Challenges in OCR of legal documents in Indian languages}
\label{sec:OCR-Challenges}

It was stated in the main text that there were various challenges in OCR of the Indian legal documents, especially the Acts in the \Actsdataset{} dataset. We briefly describe some of the OCR challenges in this section. 

\begin{figure}
\captionsetup{width=\textwidth}
\centering
\includegraphics[width=0.38\textwidth, height=4cm]{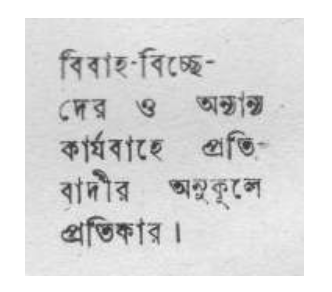}
\caption{Example of text written in an old Bengali font which is often confusing for modern OCRs}
\label{fig:Fonts}
\end{figure}

\noindent
\textbf{Presence of old Indian language fonts:}
Some Indian Act documents were scans (PDF) of very old-typed text, an example of which is shown in Figure~\ref{fig:Fonts}. The old Bengali font used in this text confused OCR systems, which have probably been trained over more recent fonts.
Further, for some Indian Languages, two or more characters may be combined to create  a compound character, and the presence of such compound characters in old fonts is even more confusing to modern OCRs.
Hence such documents that have old fonts were not considered for building the dataset.  

\begin{figure}
\captionsetup{width=\textwidth}
\centering
\includegraphics[width=0.38\textwidth, height=4cm]{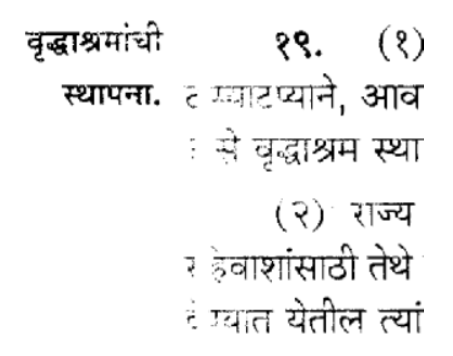}
\caption{Example of erased / eroded characters, probably due to poor quality scans. OCR has no way of giving correct output for these erased characters.}
\label{fig:Erased}
\end{figure}

\noindent
\textbf{Italicized section titles:}
Section titles in some of the documents were in italicized fonts. We found that OCRs, notably Tesseract, often cannot recognize such italicized texts correctly. However, Google Vision fared much better for such italicized texts. 

\noindent
\textbf{Presence of erased characters:}
In some documents, there were eroded / erased characters, possibly due to the poor quality of the scans. 
An example of this is shown in Figure~\ref{fig:Erased}. 
Both OCRs failed to recognize the text for such portions; hence we did not consider such portions while building the dataset (these portions were removed through manual verification).

\begin{figure*}
\centering
\captionsetup{width=\textwidth}
\includegraphics[width=\textwidth, height=2.2cm]{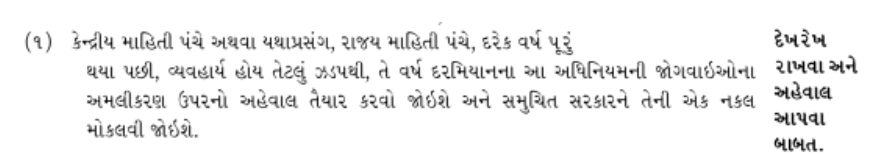}
\includegraphics[width=\textwidth, height=2.5cm]{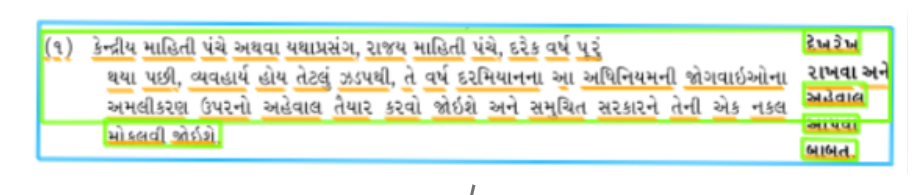}
\caption{An example of multi-column formatting affecting segmentation by OCRs. In the upper figure, the Section Text (on the left) and the Section Title (written in bold on the right) are separate, but an OCR may consider them to be part of the same sentence. 
The lower figure shows the bounding boxes of the text detected by an OCR, indicating segmentation errors and jumbled up text.}
\label{fig:Segmentation_1}
\end{figure*}

\noindent
\textbf{Complex Formatting affects text segmentation of OCR:}
An OCR system needs to recognise characters, as well as perform `segmentation' i.e., group the characters into correct units like sentences or words. 
For legal documents, the formatting can sometimes affect OCR segmentation.
An example of \textit{multi-column} format affecting text segmentation is shown in Figure~\ref{fig:Segmentation_1}. 

Thus the output of the OCR system must be carefully matched with the source document, to ensure the correctness of segmentation. 
Sometimes, we found the text blocks and sentences were jumbled up in the OCR output.
By referring to the source document, the ordering of  affected sentences / blocks were corrected manually.
Sometimes it was needed to manually crop the document into smaller regions to get better text segmentation.
All these steps made it very time-consuming to develop the \Actsdataset{} dataset. 


\section{Choice of OCR} \label{OCR Details}

Two OCR (Optical Character Recognition) systems were tried for the OCR of the \CCIdataset{} and \Actsdataset{} datasets – Tesseract \& Google Vision. 

\subsection{OCR using Tesseract}
A Python script was used to fetch PDF files, split a file into pages, apply image processing (binarization, skew removal) on each page, and then feed the page into Tesseract\footnote{\url{https://github.com/tesseract-ocr/tesseract}}.
The Tesseract output from all the pages of the file was then concatenated in order and written to a text file. 

\subsection{OCR using Google Vision}
A PDF file was split into pages, and each page of the file was uploaded to the Google Vision API\footnote{\url{https://cloud.google.com/vision/docs/samples/vision-document-text-tutorial}}. The JSON responses from the API were saved.
A Python script was designed which would take a directory containing JSON files, extract the text information from all the JSON files, and write out the combined text  to another file.
For some pages, the page had to be split into two images and processed separately due to Google Vision AI image size limits. 

\begin{figure}[tb]
\centering
\captionsetup{width=\textwidth}
\includegraphics[width=0.5\textwidth, height=7cm]{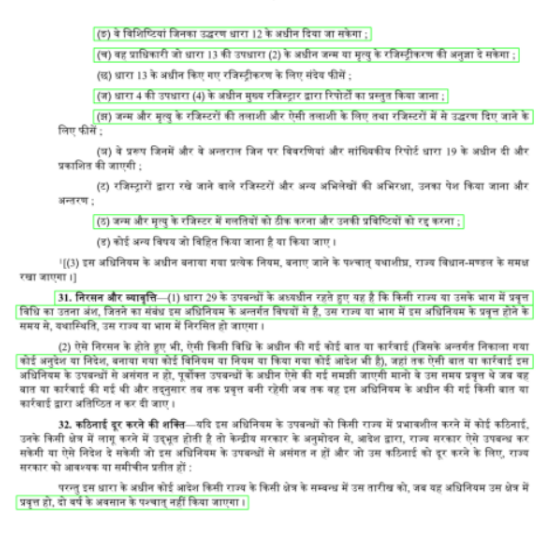}
\caption{10 randomly selected lines from a page, showing good page coverage. This is how lines were selected for comparing OCR performance.}
\label{fig:Page_Coverage}
\end{figure}

\subsection{Comparing OCR Accuracy for Google Vision and Tesseract}
A manual evaluation of 100 lines of the OCR output from the two systems was performed to assess the accuracy of text detection.
For each of the Hindi and Bengali languages, 5 pages were randomly chosen from the corpus.
From each of these pages, 10 random lines were chosen and fed to the two OCR systems.

\begin{figure}[tb]
\centering
\includegraphics[width=0.5\textwidth, height=7cm]{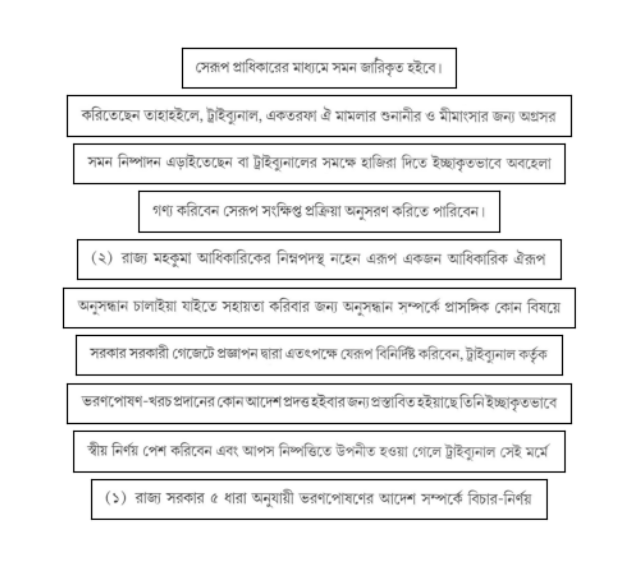}
\caption{A sample of 10 Bengali lines randomly selected from one page of a document.}
\vspace{-5mm}
\label{fig:Beng_Sample}
\end{figure}

\begin{figure}[H]
\centering
\captionsetup{width=\textwidth}
\includegraphics[width=0.6\textwidth, height=5cm]{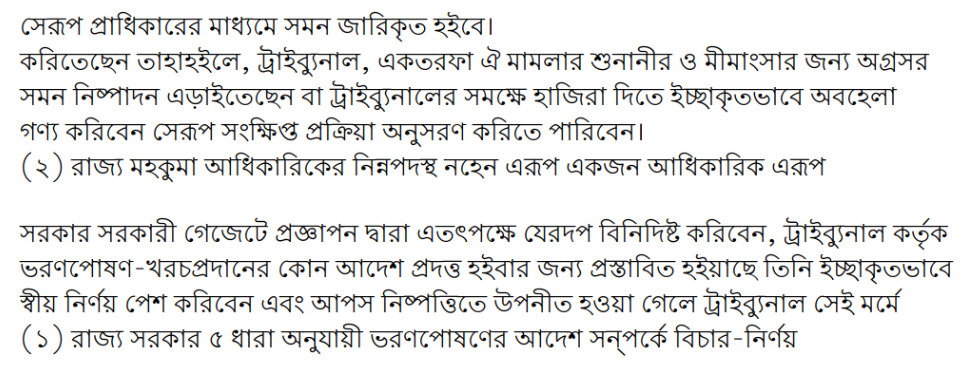}
\caption{OCR-generated text for the sample shown in Figure~\ref{fig:Beng_Sample}. For evaluating OCR performance, each character was manually matched with the respective character from the image file for each line.}
\label{fig:Beng_Sample_Text}
\end{figure}

For this evaluation, random lines of text were automatically selected from a page using Image Processing techniques.
The image was first converted into grayscale. A threshold was applied to binarize the image with the text being in white and the page being black.
After this, the image was convolved with a Gaussian Filter to smudge the text characters.
The image was then morphologically processed by applying a Dilation operation with a wide \& thin kernel.
Such a kernel was chosen to join the characters of a line together, but keep them separate from the characters of the next line. Next, contour detection was applied, to find connected regions. These contours were filtered based on some heuristics (like height, width, and position of the bounding rectangle of these contours), to obtain a list of line-contours from the image.
From this list, 10 contours were randomly chosen (which corresponded to the 10 lines). 
This method ensured good coverage of the entire page, as shown in Figure~\ref{fig:Page_Coverage}. 
These contours were then cropped from the page image and joined together to create one file.
Figure~\ref{fig:Beng_Sample} shows an example file generated from the selected lines on a page.
This file was then fed to OCR systems to get text output.
Figure~\ref{fig:Beng_Sample_Text} shows the OCR output for the sample file shown in Figure~\ref{fig:Beng_Sample}. 

\begin{table}[H]
\centering
\captionsetup{width=\textwidth}
\caption{Comparing OCR Accuracy of Tesseract \& Google Vision}
\begin{tabular}{|c|c|c|} 
 \hline
 Language & Tesseract Accuracy & Google Vision Accuracy \\
 \hline
 Bengali & 92\% & 96\% \\ 
 Hindi   & 94\% & 96\% \\
 \hline
\end{tabular}
\label{OCR_Accuracy}
\end{table}

The detected text (by the OCR) was then manually compared with the original text in the image. Each character was manually matched with the respective character from the image file for each line. 
A conservative scheme was used to identify defects -- for a particular line, if the OCR output did not match with the image for at least one character, the unit was considered defective.

The performance of the two systems, as obtained in our evalution, is given in Table~\ref{OCR_Accuracy}. 
The percentages in the table are in terms of lines of text. For a particular line, if the OCR output was not matching with the image for even one character, the unit was considered wrongly identified. 
Hence the numbers in Table~\ref{OCR_Accuracy} give the percentage of lines that were identified perfectly correctly.  

Compared to Tesseract, Google Vision had higher accuracy. 
Hence, we decided to use Google Vision for building the final datasets.